\documentclass[a4paper,fleqn]{cas-dc}

\usepackage{verbatim}
\usepackage{graphicx}
\usepackage{subcaption}         
\usepackage{adjustbox}          
\usepackage{booktabs}           
\usepackage{amsmath}            
\usepackage{afterpage}          
\usepackage{enumitem}
\usepackage{hyperref}

\usepackage[numbers]{natbib}

\begin{document}
\let\WriteBookmarks\relax
\def\floatpagepagefraction{1}
\def\textpagefraction{.001}

\shorttitle{Gravity Network for end-to-end small lesion detection}    
\shortauthors{Ciro Russo, Alessandro Bria, Claudio Marrocco}  

\title [mode = title]{Gravity Network for end-to-end small lesion detection}  


\author[1]{Ciro Russo}[orcid=0009-0002-8751-8605, bioid=1] \ead{ciro.russo@unicas.it}
\author[1]{Alessandro Bria}[orcid=0000-0002-2895-6544, bioid=2] \ead{a.bria@unicas.it}
\author[1]{Claudio Marrocco}[orcid=0000-0003-0840-7350, bioid=3]\corref{cor1} \ead{c.marrocco@unicas.it}

\cortext[cor1]{Corresponding author:  \newline Phone: +39 0776 2993381,
Fax: +39 0776 2994355}

\address[1]{Department of Electrical and Information Engineering, University of Cassino and L.M., Via G. Di Biasio 43, 03043 Cassino (FR), Italy}

\begin{abstract}
This paper introduces a novel one-stage end-to-end detector specifically designed to detect small lesions in medical images. Precise localization of small lesions presents challenges due to their appearance and the diverse contextual backgrounds in which they are found. To address this, our approach introduces a new type of pixel-based anchor that dynamically moves towards the targeted lesion for detection. We refer to this new architecture as GravityNet, and the novel anchors as gravity points since they appear to be ``attracted" by the lesions. We conducted experiments on two well-established medical problems involving small lesions to evaluate the performance of the proposed approach: microcalcifications detection in digital mammograms and microaneurysms detection in digital fundus images. Our method demonstrates promising results in effectively detecting small lesions in these medical imaging tasks.
\end{abstract}


\maketitle

\section{Introduction}\label{sec:introduction}
Detection of small lesions in medical images has emerged as a compelling area of research, which holds significant relevance in medicine, especially in fields like Radiology and Oncology, when a timely disease diagnosis is essential \citep{Loud_Murphy_2017}. Small lesions are primarily characterized by a limited size and can vary greatly in nature depending on their location and the involved tissue. In numerous real-world scenarios, the identification and classification of small lesions is a challenging and critical diagnostic process. For example, retinal microaneurysms are the earliest sign of diabetic retinopathy and are caused by small local expansion of capillaries in the retina \citep{Ezra_et_al_2013}. In ischemic stroke imaging, early identification of small occlusion is crucial to initiate timely treatment \citep{Soun_et_al_2021}. In cancer diagnosis, many forms of cancer originate as small lesions before they grow and spread, such as breast calcifications, which are one of the most important diagnostic markers of breast lesions \citep{Morgan_et_al_2005}, or pulmonary nodules, which can be the first stage of a primary lung cancer \citep{Loverdos_etl_al_2019}.
The ability to early and accurately detect small lesions can make a difference in the treatment and prognosis of patients and have a substantial impact on patient health. Manual interpretation of medical images can be time-consuming and susceptible to human error, especially when the task involves of localization and identification of small lesions within the full image space \citep{Eadie_et_al_2012}.

There is a long tradition of research on automatic lesion detection methods \citep{Shen_et_al_2017}.
Traditional image processing methods, such as thresholding, edge detection, and morphological operations, can be effective for detecting small lesions in images with clear and well-defined structures. However, these methods are limited by the presence of noise and variability in medical images. 
The use of Machine Learning, and in particular Deep Learning, helps to enhance reliability, performance, and accuracy of diagnosing systems for specific diseases \citep{Shehab_et_al_2022}. Actually,
the first lesion detection system based on Convolutional Neural Networks (CNNs) was proposed back in 1995 to detect lung nodules in X-ray images \citep{Lo_et_al_1995}.
However, only in the last ten years, CNNs have acquired great popularity thanks to their remarkable performance in computer vision \citep{LeCun_et_al_2015}, rapidly becoming the preferable solution for automated medical lesion detection \citep{Suzuki_2017, Gu_et_all_2022, Jiang_et_al_2023, Jiang_Zhou_et_al_2023}. The reason for this success is due to the ability of learning hierarchical representations directly from the images, instead of using handcrafted features based on domain-specific knowledge. CNNs are able to build features with increasing relevance, from texture to higher order features like local and global shape \citep{LeCun_et_al_1998}.

A typical CNN architecture for medical image analysis is applied to subparts of an image containing candidate lesions or background. This means that the image is divided into patches of equal size and partially overlapping, and each patch is processed individually. The output image is formed by reassembling the individually processed patches \citep{Karimi_Ward_2016}. Despite patch-based methods being widely used, they suffer from several problems, especially in the case of small lesion detection \citep{Ciga_et_al_2021}, where accurate detections requires both local information about the appearance of the lesion and global contextual information about its location. This combination is typically not possible in patch-based learning architectures \citep{Litjens_et_al_2017}, even with a multi-scale approach where the appearance of a small lesion can be missed.
An alternative is to use anchoring object detection methods of computer vision \citep{Kaur_Singh_2022}, like RetinaNet \citep{RetinaNet_2017}, which can be adapted to be used in lesion detection problems \citep{Lotter_et_al_2021}. These methods face difficulties when the objects to be detected are very small, mainly for two reasons: (i) lesions have an extremely small size compared to natural objects; (ii) lesions and non-lesions have similar appearances, making it difficult to detect them effectively \citep{Chen_et_al_2022}.

We propose a novel one-stage end-to-end detector based on a new type of anchoring technique customised to small lesion detection in medical images. Differently from classical anchor methods, which make use of anchor boxes to capture scale and aspect ratio of specific classes of objects, the proposed anchor is pixel-based and moves towards the lesion to be detected. We named \textit{GravityNet} this new architecture and \textit{gravity points} the new anchors, because they are distributed over the whole image and seem to be ``attracted" by hypothetical ``gravitational masses" located in the centres of the lesions. Such gravitational anchoring reveals to be particularly effective when small lesions have to be detected in the whole image space. To evaluate the performance of the proposed approach, we focused on two small lesions: microcalcifications on digital mammograms and microaneurysms on digital fundus images. In both cases, the lesions occupy only few pixels within an image, resulting in limited features for them to be distinguished from the surrounding tissues. Thus, their accurate localization becomes a main challenge due to their appearance and to the heterogeneity of their contextual backgrounds.

The paper is organized as follows. Section~\ref{sec:related_work} is a brief overview of object detection techniques in medical images and consequently of small lesion detection methods. Section~\ref{sec:gravity_net} introduces the proposed method. Section \ref{sec:experiments} reports the experimental analysis, followed by results in Section~\ref{sec:results}. Finally, Sections~\ref{sec:discussion} and \ref{sec:conclusions} end the paper with discussion and conclusions.

\begin{figure*}[t!]
    \centering
    \includegraphics[width=0.8\textwidth]{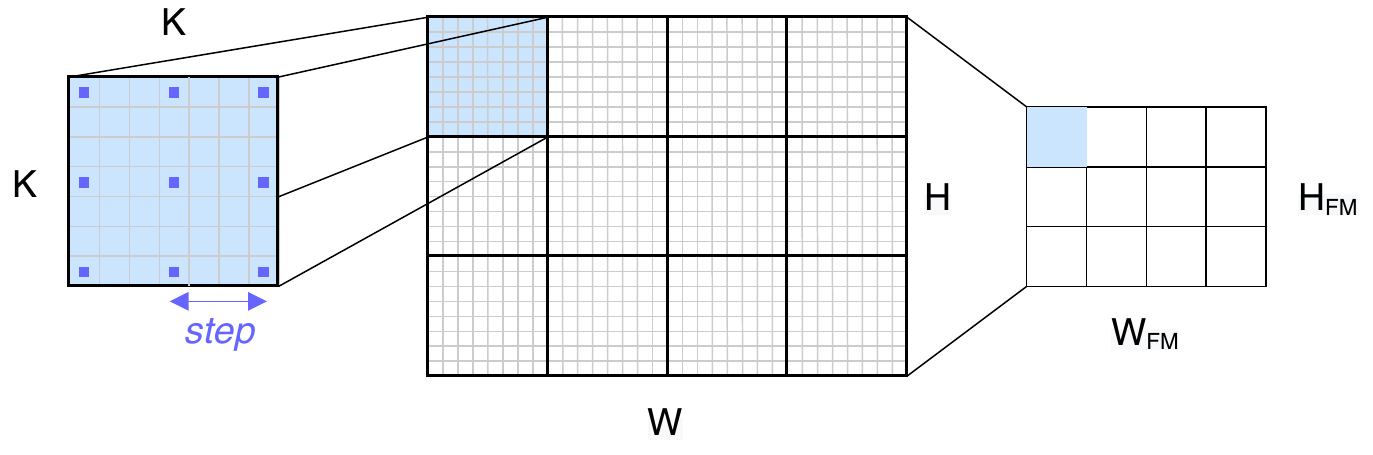}
    \caption{Gravity-points distribution: on the left, the feature grid of size $K \times K$; in the middle, the entire image $H \times W$; on the right, the feature map $H_{FM} \times W_{FM}$.}
    \label{fig:GravityPoints}
\end{figure*}

\section{Related work}\label{sec:related_work}

\subsection{Object detection in medical images}
Object detectors can be divided into two categories: (i) two-stage detector, the most representative is Faster R-CNN \citep{Faster_RCNN_2015}, (ii) one-stage detector, such as YOLO \citep{YOLO_2016}, and SSD \citep{SSD_2016}. Two-stage detectors are characterized by high localization and object recognition accuracy, whereas the one-stage detectors achieve high inference speed \citep{Jiao_et_al_2019, Zhao_et_al_2019}.
In a two-stage approach, the first stage is responsible of generating candidates that should contain objects, filtering out most of the negative proposals, whereas the second stage performs the classification into foreground/background classes and regression activities of the proposals from the previous stage.

Recently, the most popular object detection methods in computer vision have been applied to medical imaging \citep{Calli_et_all_2021, Chen_et_al_2022, Jiang_et_al_2023}. In \citep{Ding_et_all_2017}, Faster R-CNN \citep{Faster_RCNN_2015} is applied with the VGG-16 \citep{VGG16_2015} network as backbone for pulmonary nodule detection. The YOLO architecture has been modified for lymphocyte detection in immunochemistry \citep{Swiderska_et_al_2019, Rijthoven_et_al_2018} and for pneumothorax detection on chest radiographs \citep{Park_et_al_2019}. In \citep{Qiu_et_al_2022}, a deep learning algorithm based on the YOLOv5 detection model is proposed for automated detection of intracranial artery stenosis and occlusion in magnetic resonance angiography. Other studies \citep{Kim_et_al_2020, Schultheiss_et_al_2020, Lotter_et_al_2021} exploited architectures such as RetinaNet and Mask R-CNN for lung nodules and breast masses localization. In \cite{Civilibal_et_al_2023}, Mask R-CNN \citep{MaskRCNN_2017} is used by first assigning bounding boxes for each tumor volume to perform detection and classification of normal and abnormal breast tissue.

\subsection{Small lesion detection}
Although existing object detection models have been very successful with natural images \citep{Jiao_et_al_2019, Zhao_et_al_2019}, in medical images the high resolution makes the problem particularly challenging to discover small lesions, requiring complex architectures and the use of more than one stage for multi-resolution detection. In \citep{Shen_et_al_2015}, three CNN architectures, each at different scale, are applied to lung nodule detection, whereas in \citep{Kawahara_et_al_2016} a multi-stream CNN is designed to classify skin lesions, where each individual stream worked on a different image resolution.
In \citep{Wang_Yang_2018}, a context-sensitive deep neural network is developed to take into account both the local image features of a microcalcification and its surrounding tissue background. In \citep{Savelli_et_al_2020}, a multi-context architecture is proposed, based on the combination of different CNNs with variable depth and individually trained on image patches of different size. In \citep{Bria_et_al_2020}, the problem of class imbalance between lesions and background is addressed by proposing a two-stage deep learning approach where the first stage is a cascade \citep{Bria_et_al_2016} of one-level decision trees, and the second is a CNN, trained to focus on the most informative background configurations identified by the first stage.

Recently, in \citep{Wang_et_al_2022} a hierarchical deep learning framework  consisting of three models each with a specific task is proposed for bone marrow nucleated differential count. Some studies \citep{Dass_et_al_2022, Han_et_al_2022} combined image processing techniques and deep learning algorithms to evaluate lung tumor and liver tumor detection respectively. In \citep{Yurdusev_et_al_2023}, the visibility of microcalcifications in mammographic images is increased by difference filtering using the YOLOv4 model. 
A three-stage multi-scale framework for the microaneurysms detection is designed in \citep{Soares_et_al_2023}, whereas multi-scale approach based on YOLOv5 is proposed for the detection of stroke lesions \citep{Chen_Duan_et_al_2022}.
 
\begin{figure*}[t!]
    \centering
    \includegraphics[scale=.45]{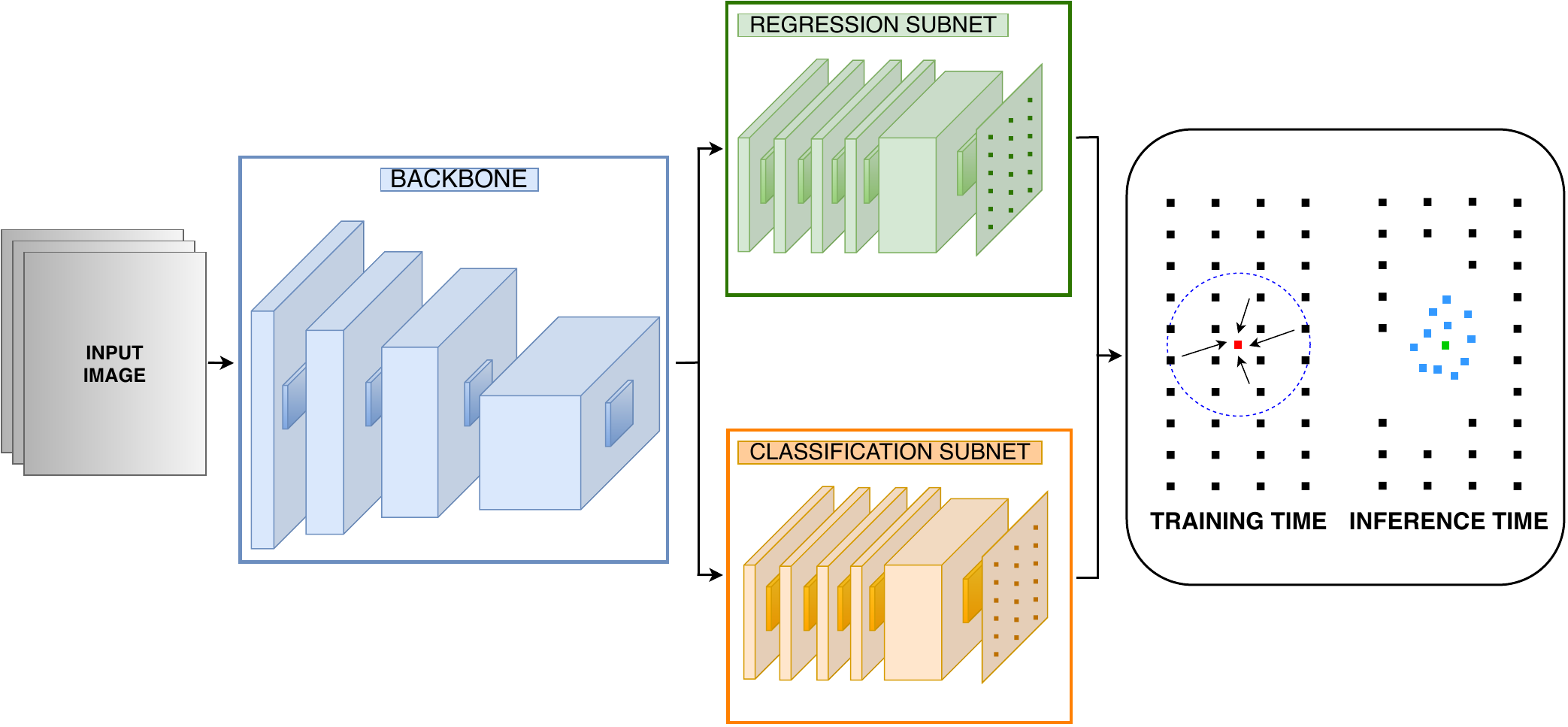}
    \caption{GravityNet architecture is composed of a backbone (blue) and two subnetworks, attached to the backbone output, one for classification task (orange) and one for regression task (green). The output is a representation of the gravity points in the grid pattern at training time and the subsequent attraction behavior towards the lesion at inference time. Gravity points in light blue correspond to positive candidates trained to collapse toward the ground truth in light green}
    \label{fig:GravityNet}
\end{figure*}

\section{GravityNet}\label{sec:gravity_net}
This section explains the proposed network architecture and the concept of \textit{gravity points}, a new anchoring technique designed for small lesion detection.

The code is available at this \href{https://github.com/cirorusso2910/GravityNet}{link}.

\subsection{Gravity points}\label{sec:GravityPoints}
We define a \textit{gravity point} as a pixel-based anchor, which inspects its surroundings to detect lesions.
The gravity-points distribution is generated with a grid of points spaced by a user-defined \textit{step} parameter. A base configuration is generated in a squared reference window, named \textit{feature grid}, of size $K \times K$ where $K$ is equal to the upper integer of the ratio between the dimensions of the image and the feature map:
\begin{equation}\label{Eq:FeatureGrid}
    K \times K = \Biggl\lceil \frac{H}{H_{FM}} \Biggl\rceil \text{ } \times \text{ } \Biggl\lceil \frac{W}{W_{FM}} \Biggl\rceil
\end{equation}
Assuming that the first gravity point is located in the upper left corner of the feature map, the number of gravity points in a feature grid is equal to: 

\begin{equation}\label{Eq:NumGravityPointsFeautureGrid}
    N^{FG}_{GP} = \Biggl( \text{ } \Biggl\lfloor \frac{K-2}{step} \Biggl\rfloor \text{ }  + 1 \Biggr)^2
\end{equation}
where $0 < step \leq K - 2$. In cases where $K - 2$ is multiple of \textit{step} the distribution will be equispaced in the feature grid.

Since each pixel in the feature map corresponds to a feature grid in the image, the complete configuration is obtained by sliding the base configuration over the whole image. The total number of gravity points $N_{GP}$ in the image is equal to the base configuration times the number of feature grids:
\begin{equation}\label{Eq:NumGravityPoints}
    N_{GP} = N^{FG}_{GP} \cdot H_{FM} \cdot W_{FM}
\end{equation}
Fig.~\ref{fig:GravityPoints} shows an example of gravity-points distribution.

\subsection{Architecture}\label{sec:Architecture}
GravityNet is a one-stage end-to-end detector composed of a backbone network and two specific subnetworks. The backbone is a convolutional network and plays the role of feature extractor. The first subnet performs convolutional object classification on the backbone output, whereas the second subnet performs convolutional gravity-points regression. Fig.~\ref{fig:GravityNet} shows the overall architecture.

The backbone is the underlying network architecture of a detection model and provides a feature map containing basic features and representations of input data, which are then processed to perform a specific task. The bottom layers of a backbone net usually extract simple features such as edges and corners, while the top layers learn more complex features like parts of lesions. The feature maps generated by these layers are used as a representation of the input image and fed into two models for classification and regression tasks.

The classification subnet is a fully convolutional network that outputs the probability of lesion presence at each gravity-point location. The subnetwork applies four $3 \times 3$ convolutional layers, each with $256$ filters, where the first one maps the number of features output from the backbone, followed by ReLU activations. The last layer applies a filter with $N_{AP} \cdot 2$ outputs and sigmoid activation to obtain the binary predictions for each gravity point.

The regression subnetwork is connected to the output of the backbone with the purpose of regressing the offset from each gravity point to the closest lesion. The design of the regression subnet is the same of the classification subnet. The last layer outputs $N_{AP} \cdot 2$ values, indicating the offsets to move each gravity point towards a lesion.
It is worth noting that the classification and regression subnets, though sharing a common structure, use separate parameters.

\subsection{Gravity loss}\label{sec:FocalLoss}
\textit{Gravity Loss} (GL) is a multi-task loss that contains two terms: one for regression (denoted as $GL_{reg}$) and the other for classification (denoted as $GL_{cls}$).

The multi-task loss can be written as:

\begin{equation}\label{Eq:GravityLoss}
    GL = GL_{cls} + \lambda GL_{reg}
\end{equation}
where $\lambda$ is an hyperparameter that controls the balance between the two task losses.

\begin{figure}[t!]
    \centering
    \includegraphics[scale=0.7]{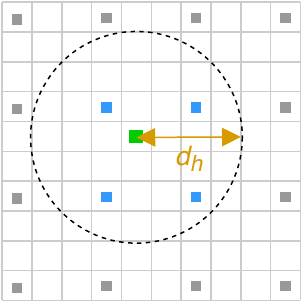}
    \caption{Hooking process where gravity points (light blue) are hooked to a lesion (light green)}
    \label{fig:GravityPointsHooking}
\end{figure}

\subsubsection{Classification loss}\label{subsec:ClassificationLoss}
 Since significant class imbalance between lesion and background is usually present in medical images \citep{Bria_et_al_2020}, the classification loss is a variant of Focal Loss \citep{RetinaNet_2017}. This loss is designed to address the issue of class imbalance in object detection tasks, where the majority of the examples belong to the negative class (e.g., background) and only a few examples belong to the positive class (e.g., lesion).

The classification loss is defined as:
\begin{equation}
    GL_{cls} = -\alpha_t \cdot (1 - p_t)^\varphi \cdot \log(p_t)
\end{equation}
where $p_t$ is the predicted probability of the true class (lesion), $\varphi$ is a focusing parameter that controls the rate at which the modulating factor $\alpha_t$ decreases as the predicted probability $p_t$ increases.

To evaluate $p_t$ with gravity points, we introduce a criterion based on the Euclidean distance between the gravity points and the ground-truth lesions\footnote{Without loss of generality, we consider as ground truth the center of the smallest bounding box containing the lesion.}. We consider as belonging to the positive class those gravity points with a distance from the closest ground-truth lesion lower than a threshold distance that we named \textit{hooking distance} $d_h$. All the gravity points within that distance are hooked to the lesion and trained to move towards it. Fig.~\ref{fig:GravityPointsHooking} shows an example of gravity-points hooking process.

\begin{figure}[!t]
    \centering
    \vspace{0.3cm}
    \includegraphics[width=0.35\textwidth]{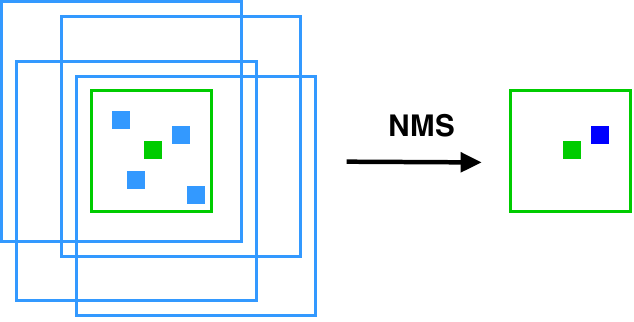}
    \caption{An example of NMS: on the left, gravity points and corresponding boxes (light blue) hooked to a lesion (green); on the right, the final candidate corresponding to the gravity point with the highest score (blue)}
    \label{fig:NMS}
\end{figure}

\subsubsection{Regression loss}\label{subsec:RegressionLoss}
Let us indicate as $(d_x, d_y)$ the distance between a gravity point and the relative hooked lesion, and as $(o_x, o_y)$ the output of the regression subnetwork, which represents the offset to move each gravity point towards the hooked lesion.

We evaluate the regression loss as:
\begin{equation}
    GL_{reg} = \sum_{\forall\text{ }hooked\text{ }GP} \sum_{i \in \{x, y\}} smooth_{L1}(d_i - o_i)
\end{equation}
where $smooth_{L1}(t)$ is the Smooth L1 loss \citep{Faster_RCNN_2015}, defined as:
\begin{equation}
    smooth_{L1}(t) =
        \begin{cases}
            0.5t^2, & \text{if } |t| < 1 \\
            |t| - 0.5, & \text{otherwise}
        \end{cases}
\end{equation}

\subsection{Inference time}\label{subsec:InferenceTime}
The model produces two output predictions for each gravity point for each subnetwork. \textit{Non-Maxima-Suppression} (NMS) is applied to reduce the number of false candidates (see Fig.~\ref{fig:NMS}): (i) an $L \times L$ box is built for each hooked gravity points, where $L$ is chosen equal to the average size of the lesions to be detected; (ii) all boxes with an Intersection over Union (IoU) greater than 0.5 are merged; and (iii) for each merger, the gravity point with the highest score is considered as final candidate.
After NMS, we determine the lesion class with a threshold $\gamma$ on the classification score: all predictions with scores above $\gamma$ belong to the positive class (lesion), the remaining ones to the negative class (no lesion).

\section{Experiments}\label{sec:experiments}
We proved the effectiveness of \textit{GravityNet} on two detection problems in medical image analysis: (i) microcalcifications on full field digital mammograms and (ii) microaneurysms on digital ocular fundus images.

Microcalcifications (MCs) are calcium deposits and are considered as robust markers of breast cancer \citep{Logullo_et_al_2022}. MCs appear as fine, white specks, similar to grains of salt, with size between $0.1$ $mm$ and $1$ $mm$. Due their small dimensions and the inhomogeneity of the surrounding breast tissue, identifying MCs is a very challenging task. Moreover, mammograms contain a variety of linear structures (such as vessels, ducts, etc.) that are very similar to MCs in size and shape, making detection even more complex.

Microaneurysms (MAs) are the earliest visible manifestation of Diabetic Retinopathy, one of the leading causes of vision loss globally \citep{Lee_et_al_2015}. MAs are described as isolated small red dots of $10$-$100$ $\mu m$ of diameter sparse in retinal fundus images, but sometimes they appear in combination with vessels. Retinal vessels, together with dot-hemorrhages and other objects like the small and round spots resulting from the crossing of thin blood vessels, make MAs hard to distinguish.

\subsection{Dataset}\label{subsec:dataset}

\subsubsection{Microcalcifications dataset}
We used the publicly available INBreast database \citep{INbreast_2012}, acquired at the Breast Centre in Centro Hospitalar de S. Jo\~ ao (CHSJ) in Porto, Portugal. The acquisition equipment was the MammoNovation Siemens FFDM, with a solid-state detector of amorphous selenium, pixel size of $70$ $\mu m$ (microns) and 14-bit contrast resolution.
The image matrix was $4,084 \times 3,328$ ($243$ images) or $3,328 \times 2,560$ ($167$ images), depending on the compression plate used in the acquisition and according to the breast size of the patient.

The database has a total of $410$ images, amounting to 115 cases, from which $90$ cases are from women with both breasts, and $25$ are from mastectomy patients.
Calcifications can be found in $313$ images for a total of $7,142$ individual calcifications. In this work, only calcifications with a radius of less than $7$ pixels were considered for testing, for a total of $5,657$ microcalcifications identified in $296$ images.

Mammograms have been cropped to the size $3,328 \times 2,560$ to have all images in the dataset with equal size. We ensured that no MC was missed after cropping.

\subsubsection{Microaneurysms dataset}
We used the publicly available E-ophtha database \citep{Eophtha_2013}, designed for scientific research in Diabetic Retinopathy. The acquired images have dimensions ranging from $960 \times 1,440$ to $1,696 \times 2,544$ with a $45^\circ$ field of view (FOV) and a pixel size of 7-15 $\mu m$. The database has a total of $381$ images: $148$ images from unhealthy patients containing $1,306$ microaneurysms, and $233$ images from healthy patients. 

The original retinal fundus images are RGB, but in this work, green channel has been extracted due to its rich information and high contrast in comparison with the other two color channels \citep{Tsiknakis_et_al_2021}. We also evaluated the average dimensions of all the retinas in the dataset and  resized all the images to an average dimensions of $1,216 \times 1,408$.

\begin{table}[h!]
\caption{Data overview}
\vspace{0.3cm}
\centering
\begin{adjustbox}{scale=0.75}
    \begin{tabular}{l  c  c  c  c  c  c  c  c}
            \hline
            \\ [0.3pt]
            \rule{0pt}{10pt} \textbf{INbreast} & \multicolumn{2}{c}{Images} & & \multicolumn{2}{c}{Unhealthy} & & \multicolumn{2}{c}{MCs} \\ [1pt]
            \cmidrule(lr){2-3} \cmidrule(lr){5-6} \cmidrule(lr){8-9}
            & 1-fold & 2-fold & & 1-fold & 2-fold & & 1-fold & 2-fold \\ [1pt]
            \hline
            \rule{0pt}{15pt} Train       & 143 & 143  & & 108 & 117 & & 2,408 & 2,051   \\ [1pt]
            \rule{0pt}{15pt} Validation  &  62 &  62  & & 39  & 45  & &  516  & 724    \\ [1pt]
            \rule{0pt}{15pt} Test        & 205 & 205  & & 154 & 142 & & 2,756 & 2,901   \\ [1pt]
            \\ [0.3pt]
            \hline
            \hline
            \\ [0.3pt]
            \rule{0pt}{10pt} \textbf{E-ophtha-MA} & \multicolumn{2}{c}{Images} & & \multicolumn{2}{c}{Unhealthy} & & \multicolumn{2}{c}{MAs} \\ [1pt]
            \cmidrule(lr){2-3} \cmidrule(lr){5-6} \cmidrule(lr){8-9}
            & 1-fold & 2-fold & & 1-fold & 2-fold & & 1-fold & 2-fold \\ [1pt]
            \hline
            \rule{0pt}{15pt} Train       & 154 & 151 & & 60 & 60  & & 542 & 552   \\ [1pt]
            \rule{0pt}{15pt} Validation  &  38 &  38 & & 14 & 14  & & 105 & 107   \\ [1pt]
            \rule{0pt}{15pt} Test        & 189 & 192 & & 74 & 74  & & 659 & 647   \\ [1pt]
            \hline
        \label{tab:DataOverview}
    \end{tabular}
\end{adjustbox}
\end{table}

\subsection{Data preparation}
For all experiments we applied 2-fold image-based cross-validation. The dataset is divided into two equal-sized folds, where one fold is used as training set and the other as test set and vice versa. A subset of the training set fold is used as validation set for parameter optimisation. See Tab.~\ref{tab:DataOverview} for more details.

In both datasets, data augmentation techniques are used to address class imbalance and enhance the model robustness and accuracy \citep{Kisantal_et_al_2019}. Three more samples for each image are generated by using horizontal and vertical flipping. All data are normalized with min-max transformation.

\begin{figure*}[h!]
    \centering
    \begin{tabular}{cccccc}
        \includegraphics[width=0.17\textwidth]{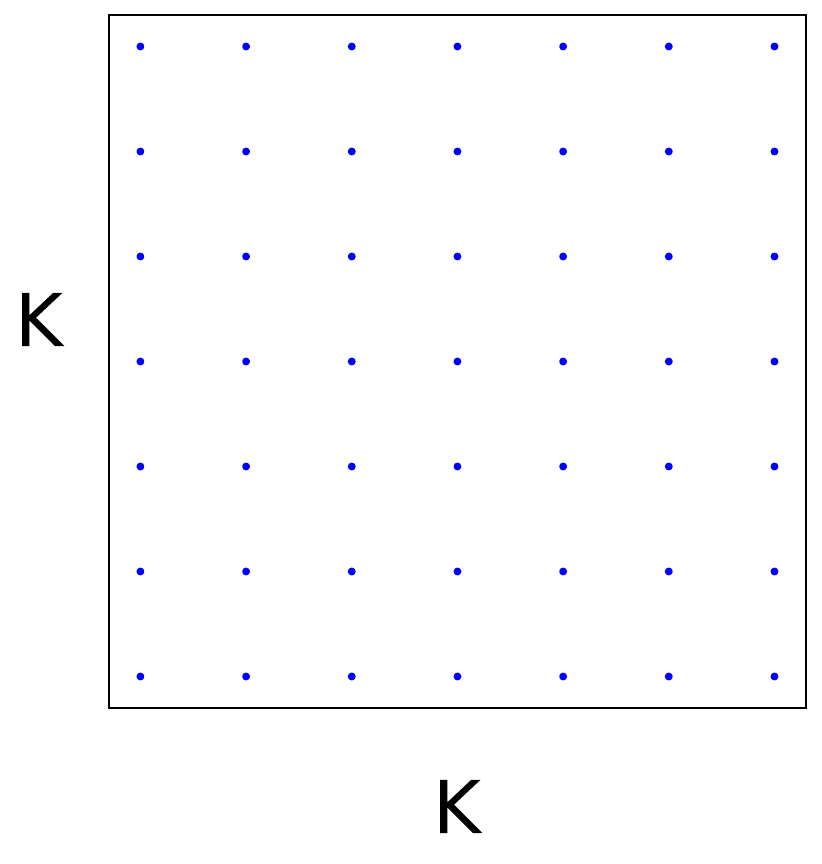} &
        \includegraphics[width=0.17\textwidth]{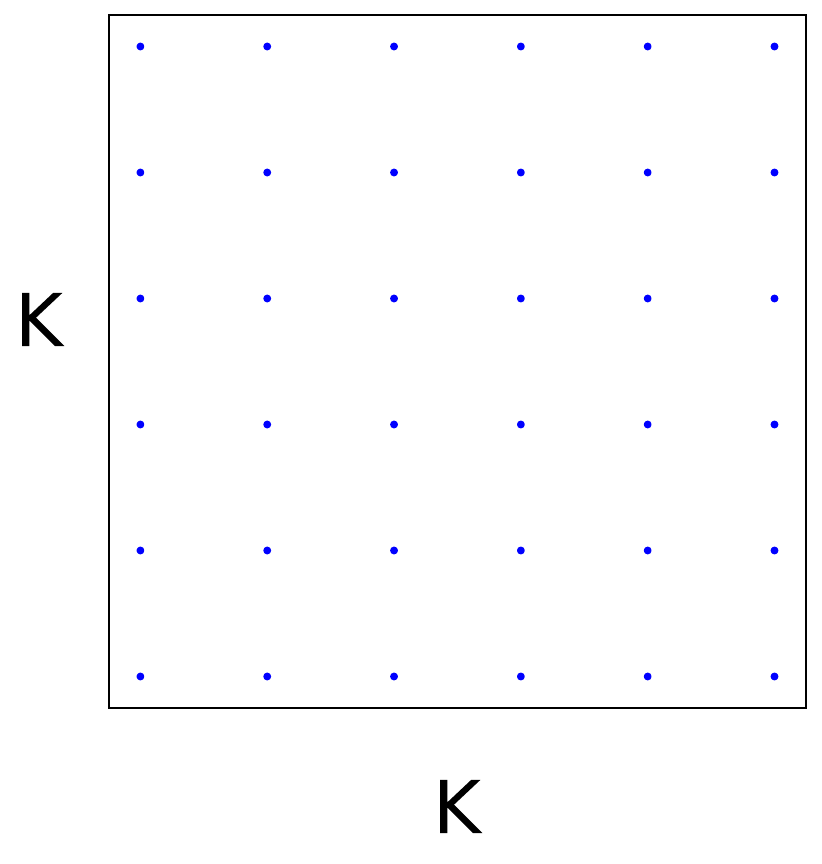} &
        \includegraphics[width=0.17\textwidth]{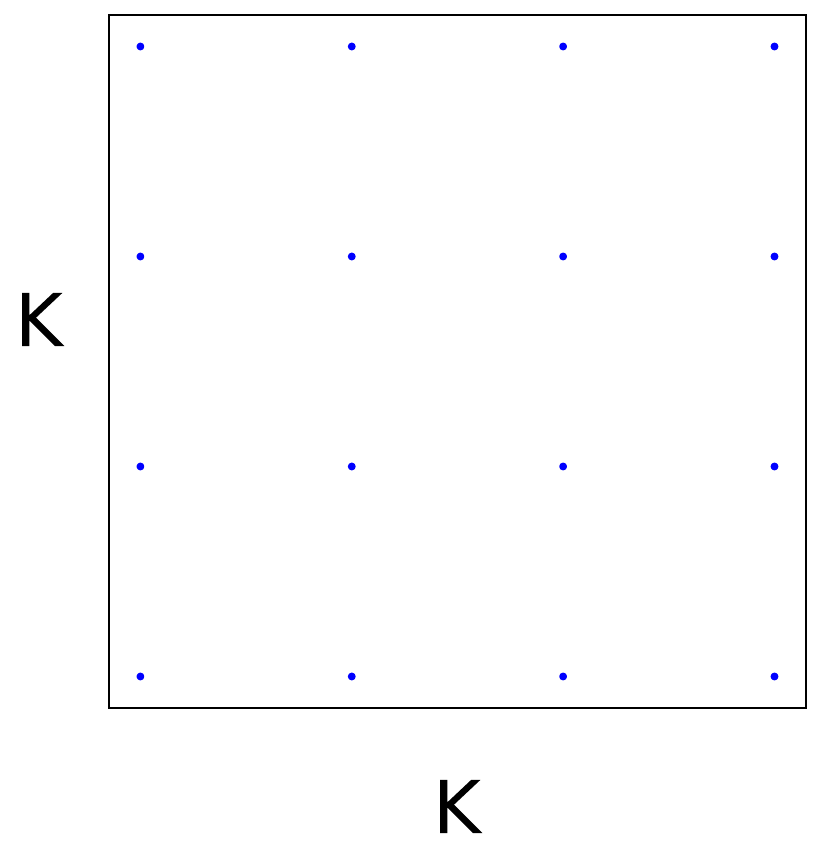} &
        \includegraphics[width=0.17\textwidth]{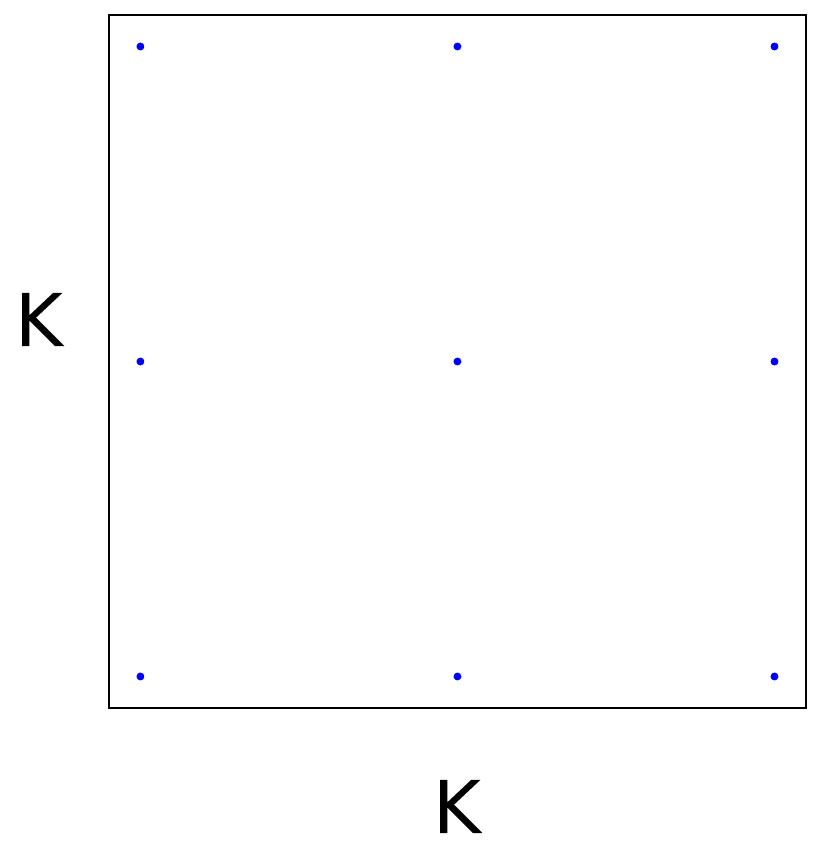} &
        \includegraphics[width=0.17\textwidth]{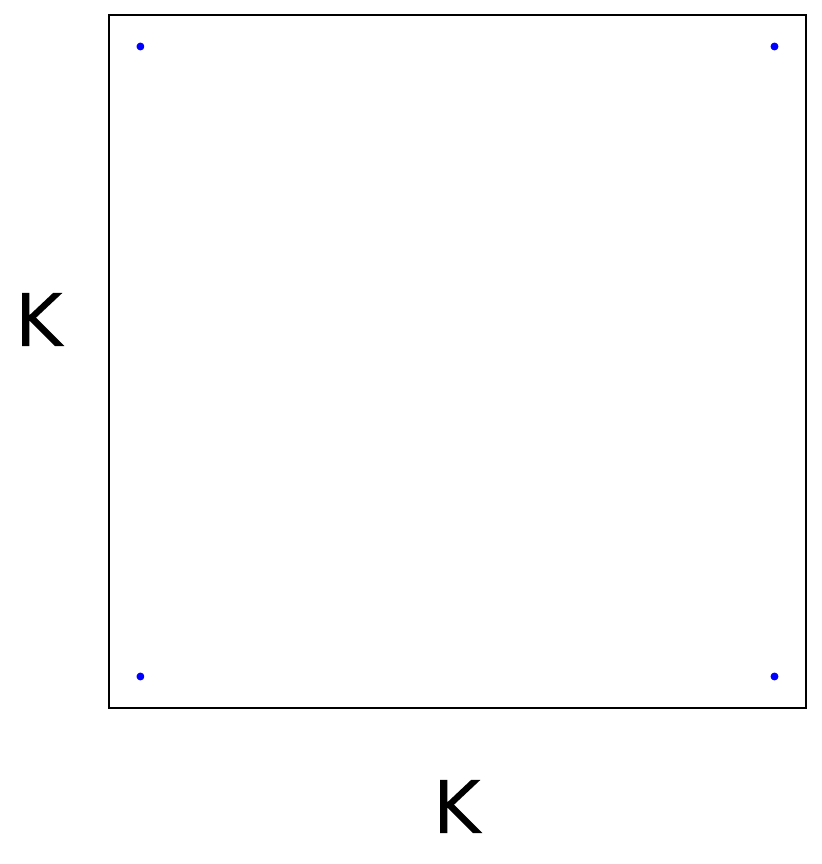} \\
        \textbf{(a)} & \textbf{(b)} & \textbf{(c)} & \textbf{(d)} & \textbf{(e)}  \\[6pt]
    \end{tabular}
    \caption{Examples of initial gravity-points configurations represented in a reference window $K \times K$, where: \textbf{(a)} \textit{step} = 5 \textbf{(b)} \textit{step} = 6 \textbf{(c)} \textit{step} = 10 \textbf{(d)} \textit{step} = 15 \textbf{(e)} \textit{step} = 30}
\label{fig:GravityPointsInitialDistribution}
\end{figure*}

\subsection{Architecture parameters}
GravityNet uses ResNet \citep{ResNet_2015} as its backbone, in order to solve the well-known vanishing/exploding gradient problems \citep{Bengio_Simard_Frasconi_1994} by using residual connections. ResNet is composed of $5$ max-pooling layers, each halving the dimensions of the feature map. According to the input dimensions, the feature map size is $104 \times 80$ for mammograms and $38 \times 44$ for retina fundus images.
As a consequence, according to Eq.~\ref{Eq:FeatureGrid}, we obtain $K=32$ and a feature grid of $32 \times 32$. We generate gravity-points configurations with \textit{step} multiple of $K - 2$ to ensure equi-spatiality. To take into account the computational cost, we chose to use configurations that did not exceed $300,000$ gravity points. Fig.~\ref{fig:GravityPointsInitialDistribution} shows some examples of the initial configurations used in this work. 
To ensure that at least one gravity point hooks a lesion, the \textit{hooking distance} $d_h$ was always chosen equal to the \textit{step}.
At inference time, we use NMS with $L=7$ for MCs and $L=3$ for MAs, which correspond to the average size of the lesions to be detected.

We train ResNet in transfer learning by using a model pretrained on natural images \citep{Yu_et_al_2022}. Both subnetworks are initialised with Xavier technique \citep{Xavier_2010}. During training we apply Adam optimization algorithm \citep{Adam_2017}. The learning rate was set to an initial value of $10^{-4}$ and decreased with a factor of $0.1$ with \textit{patience} equal to $3$. The balance between the two task losses (see Eq.~\ref{Eq:GravityLoss}) is handled by $\lambda$ equal to $10$. The batch size is by default equal to $8$. All training parameters were optimized on the validation set. The training was stopped after $50$ epochs. Experiments were conducted on a GPU NVIDIA A100 80GB.

\subsection{FROC analysis}
The detection quality was evaluated in terms of lesion-based Free-Response Operating Characteristic (FROC) curve by plotting True Positive Rate (TPR) against the average number of False Positives per Image (FPpI) for a series of thresholds $\gamma$ on the classification score associated to each sample.

A prediction with a value higher than $\theta$ is considered as True Positive (TP) when its distance from the center of a lesion is no larger than the largest side of the bounding box containing the ground-truth lesion. Otherwise, it is considered as False Positive. Notably, (i) if multiple predictions are associated to the same lesion, only the one with the highest classification score is selected as TP, and (ii) all predictions for gravity points outside the tissue were ignored.

To analyze and compare FROC curves, we chose the non-parametric approach suggested in \citep{Chakraborty_2008}. The figure-of-merit is the Partial Area under the FROC curve ($AUFC_\gamma$) to the left of $FPpI = \gamma$ calculated by trapezoidal integration. We normalized $AUFC_\gamma$ by dividing with $\gamma$ to obtain an index in the range $[0, 1]$. In particular, for both MCs and MAs detection, we selected $\gamma = 10$, a commonly used value in the literature of the respective fields \citep{Dashtbozorg_et_al_2018, Savelli_et_al_2020}. All results are presented in percentage values.

\section{Results}\label{sec:results}

\subsection{Model analysis}
To verify the effectiveness of the model, for both small lesion detection problems, experiments were conducted using different gravity-points configurations for all different depths of ResNet\footnote{It is worth noting that, due to memory constraints, for MCs detection, we use in training a batch size equal to $4$ for \textit{ResNet-50} and $2$ for \textit{ResNet-101} and \textit{ResNet-152}}. Results are reported in Tab.~\ref{tab:ResultsMCs} for MCs, and in Tab.~\ref{tab:ResultsMAs} for MAs together with the parameters of the gravity-points configurations. The best result for each backbone is shown in bold, whereas the best of all in italic. FROC curves of the best ResNet configurations are shown in Fig.~\ref{fig:FROC_Results}.

For MCs, the best result is a $AUFC_\gamma$ equal to $72.25 \%$ by using \textit{ResNet-34} and \textit{step} $10$. Configuration with \textit{step} $10$ turns to be the best for all backbones, except \textit{ResNet-50}, which achieves a $AUFC_\gamma$ equal to $71.25 \%$ with \textit{step} $6$. Dense configurations present better results with shallower backbones, e.g. with 
a \textit{ResNet-18} we obtain a $AUFC_\gamma$ equal to $70.89 \%$ and $71.47 \%$ respectively with \textit{step} $6$ and $10$ as opposed to $65.58 \%$ and $55.90\%$ respectively with \textit{step} $15$ and $30$.

For MAs, the highest $AUFC_\gamma$ ($71.53 \%$) is obtained with a \textit{ResNet-50} and \textit{step} $6$. Configuration with \textit{step} $6$ turns to be the best for all backbones, except \textit{ResNet-18}, which achieves a $AUFC_\gamma$ equal to $65.36 \%$ with \textit{step} $10$. Sparse configurations decrease the performance, even with deeper backbones, e.g. with a \textit{ResNet-152} we obtain a $AUFC_\gamma$ of $67.51 \%$ and $54.18 \%$ respectively with \textit{step} $15$ and $30$ as opposed to $65.81 \%$ and $69.86 \%$ respectively with \textit{step} $5$ and $6$.

Through result analysis, it becomes evident that we need to find the appropriate density configuration for addressing the detection problem at hand. A sparse configuration might fail to identify all lesions, particularly in the case of small ones, whereas a dense configuration could potentially generate a high number of lesion candidates.

\afterpage{
    \begin{table*}[t!]
    \caption{Results for MCs detection in terms of \% of $AUFC_\gamma$}
    \vspace{0.3cm}
    \centering
    \begin{adjustbox}{scale=0.9}
        \begin{tabular}{c  c  c  c  c  c  c  c}
                \hline
                \rule{0pt}{10pt} \textbf{Backbone} & \multicolumn{4}{c}{\textbf{Configuration}} \\ [1pt]
                \cmidrule(lr){2-5}
                \rule{0pt}{10pt} & \textit{$step=6$} \hspace{0.1cm} \textit{$d_h=6$} \hspace{0.5cm} & \textit{$step=10$} \hspace{0.1cm} \textit{$d_h=10$} \hspace{0.5cm} & \textit{$step=15$} \hspace{0.1cm} \textit{$d_h=15$} \hspace{0.5cm} & \textit{$step=30$} \hspace{0.1cm} \textit{$d_h=30$} \\ [1pt]
                \rule{0pt}{10pt} & \textit{$N_{GP}$=299,520} \hspace{0.5cm} & \textit{$N_{GP}$=133,120} \hspace{0.5cm} & \textit{$N_{GP}$=74,880} \hspace{0.5cm} & \textit{$N_{GP}$=33,280} \\ [1pt]
                \hline
                \rule{0pt}{10pt} ResNet-18  & 70.89 & \textbf{71.47} & 65.58 & 55.90 \\ [2pt]
                \rule{0pt}{10pt} ResNet-34  & 65.08 & \textbf{\textit{72.25}} & 67.44 & 56.17 \\ [2pt]
                \rule{0pt}{10pt} ResNet-50  & \textbf{71.25} & 67.73 & 69.31 & 57.12 \\ [2pt]
                \rule{0pt}{10pt} ResNet-101 & 58.85 & \textbf{64.90} & 41.69 & 53.05 \\ [2pt]
                \rule{0pt}{10pt} ResNet-152 & 60.60 & \textbf{64.86} & 62.98 & 53.86 \\ [2pt]
                \hline
            \label{tab:ResultsMCs}
        \end{tabular}
    \end{adjustbox}
    \end{table*}
    
    \begin{table*}[t!]
    \caption{Results for MAs detection in terms of \% of $AUFC_\gamma$}
    \vspace{0.3cm}
    \centering
    \begin{adjustbox}{scale=0.9}
        \begin{tabular}{c  c  c  c  c  c  c  c}
                \hline
                \rule{0pt}{10pt} \textbf{Backbone} & \multicolumn{5}{c}{\textbf{Configuration}} \\ [1pt]
                \cmidrule(lr){2-6}
                \rule{0pt}{10pt} & \textit{$step=5$} \hspace{0.1cm} \textit{$d_h=5$} \hspace{0.5cm} & \textit{$step=6$} \hspace{0.1cm} \textit{$d_h=6$} \hspace{0.5cm} & \textit{$step=10$} \hspace{0.1cm} \textit{$d_h=10$} \hspace{0.5cm} & \textit{$step=15$} \hspace{0.1cm} \textit{$d_h=15$} \hspace{0.5cm} & \textit{$step=30$} \hspace{0.1cm} \textit{$d_h=30$} \\ [1pt]
                \rule{0pt}{10pt} & \textit{$N_{GP}$=81,928} \hspace{0.5cm} & \textit{$N_{GP}$=60,192} \hspace{0.5cm} & \textit{$N_{GP}$=26,752} \hspace{0.5cm} & \textit{$N_{GP}$=15,048} \hspace{0.5cm} & \textit{$N_{GP}$=6,688} \\ [1pt]
                \hline
                \rule{0pt}{10pt} ResNet-18  & 60.95 & 61.42 & \textbf{65.36} & 63.17 & 53.88 \\ [2pt]
                \rule{0pt}{10pt} ResNet-34  & 65.01 & \textbf{68.57} & 68.38 & 64.80 & 58.76 \\ [2pt]
                \rule{0pt}{10pt} ResNet-50  & 68.89 & \textbf{\textit{71.53}} & 64.57 & 68.25 & 54.33 \\ [2pt]
                \rule{0pt}{10pt} ResNet-101 & 66.07 & \textbf{69.77} & 69.13 & 65.59 & 54.97 \\ [2pt]
                \rule{0pt}{10pt} ResNet-152 & 65.81 & \textbf{69.86} & 66.84 & 67.51 & 54.18 \\ [2pt]
            \hline
            \label{tab:ResultsMAs}
        \end{tabular}
    \end{adjustbox}
    \end{table*}
    
    \begin{figure*}[t!]
    \centering
    \begin{subfigure}[t]{0.45\textwidth}
        \includegraphics[width=\linewidth]{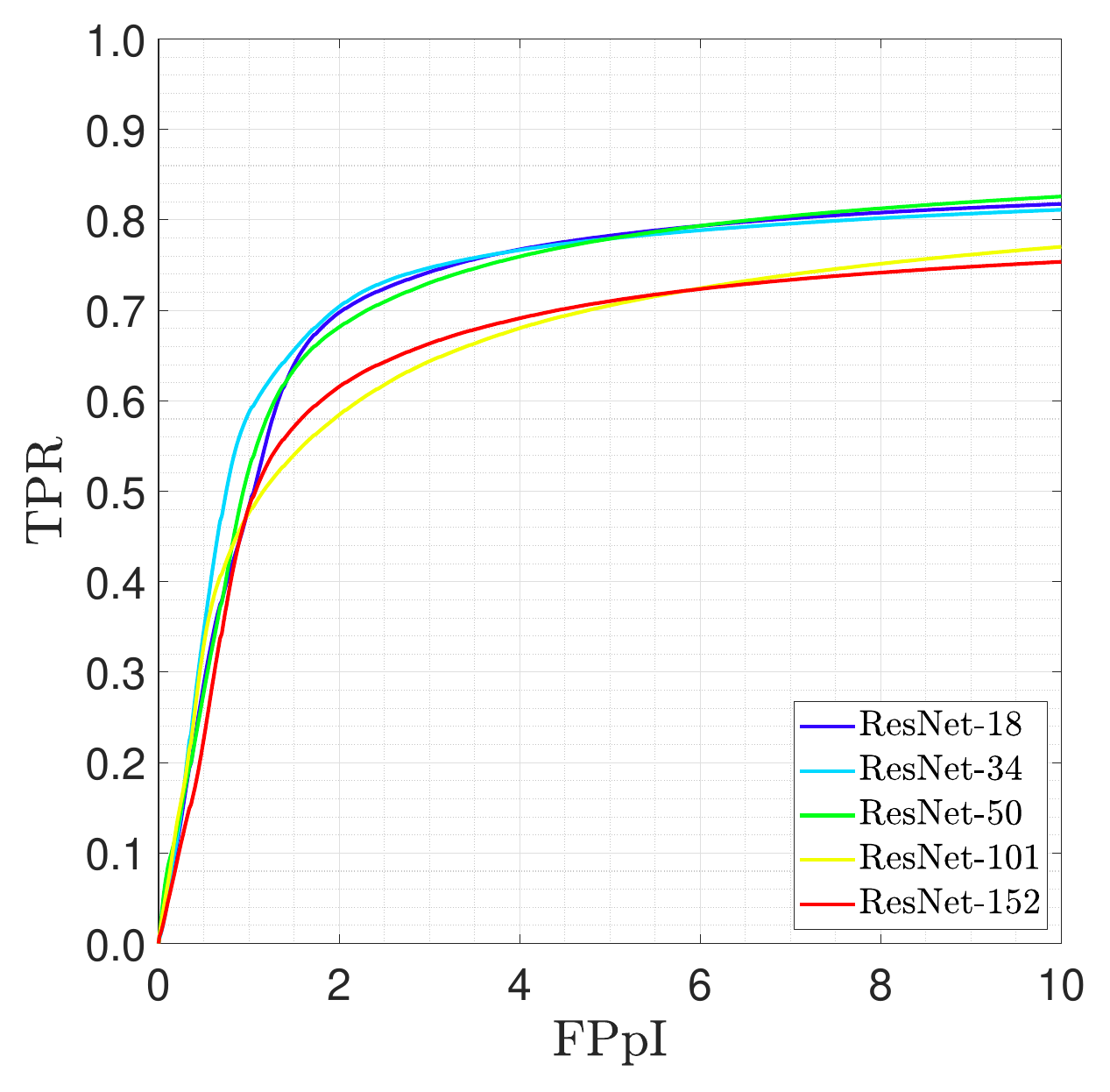}
        \caption{}
        \label{subfig:FROC_Results_MCs}
    \end{subfigure}\hspace{0.3cm} 
    \begin{subfigure}[t]{0.45\textwidth}
        \includegraphics[width=\linewidth]{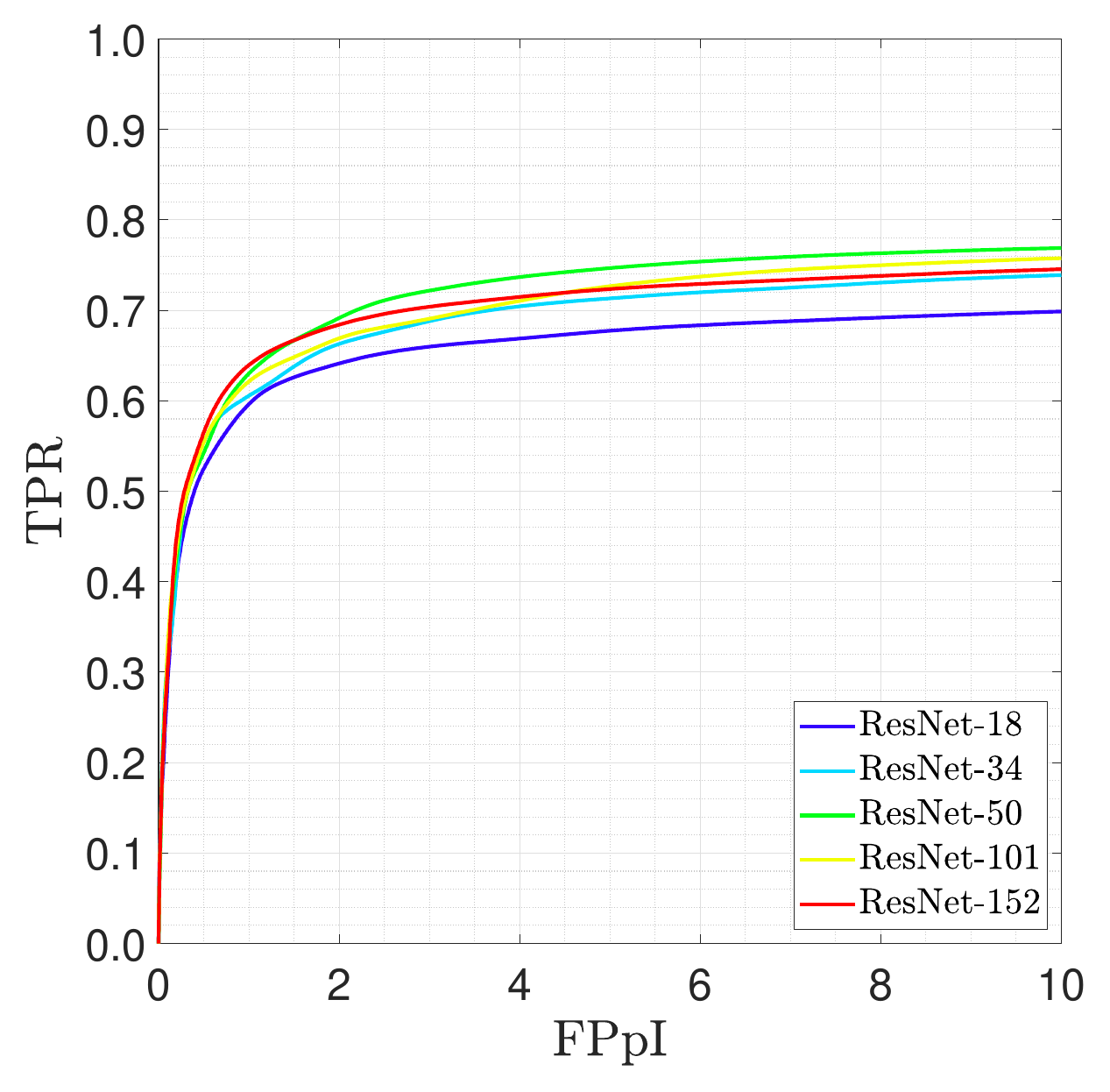}
        \caption{}
        \label{subfig:FROC_Results_MAs}
    \end{subfigure}
    \caption{FROC results with the best gravity-points configurations for each ResNet backbone on INbreast (a) and E-ophtha-MA (b)}
    \label{fig:FROC_Results}
    \end{figure*}
\clearpage\clearpage
}

\subsection{Comparison with the literature}
We compare our best models, that are \textit{ResNet-34} with \textit{step} $10$ for MCs detection and \textit{ResNet-50} with \textit{step} $6$ for MAs detection, with methods proposed in the scientific literature for the detection problems at hand:
\begin{itemize}
    \item [-] Context-Sensitive CNN (CSNet) \citep{Wang_Yang_2018}: the architecture comprises two convolutional subnetworks: one for processing the large image context with a window of size $96 \times 96$ pixels and another for processing the small microcalcification texture with a window of size $9 \times 9$ pixels. The features extracted from both subnetworks are subsequently merged and fed into a fully connected network.
    \item [-] Deep Cascade (DC) \citep{Bria_et_al_2016}: a cascade of decision stumps able to learn effectively from heavily class-unbalanced datasets. It builds on Haar features computed in a small detection window of $12 \times 12$ pixels, which can contain diagnostically relevant lesions, while limiting the exponential growth of the number of features that are extracted during training. 
    \item [-] MCNet with DC hard mining (DC-MCNet) \citep{Bria_et_al_2020}: a two-stage patch-based deep learning framework, which comprises a DC for hard mining the background samples, followed by a second stage represented by a CNN that discriminates between lesions and the more challenging background configurations.
    \item [-] Multicontext Ensemble of MCNets (ME-MCNet) \citep{Savelli_et_al_2020}: a multi-context ensemble of CNNs aiming to learn different levels of image spatial context by training multiple-depth networks on image patches of different dimensions ($12 \times 12$, $24 \times 24$, $48 \times 48$, and $96 \times 96$).
\end{itemize}

To evaluate the behavior of the proposed anchoring technique, we also compared with RetinaNet \citep{RetinaNet_2017}, a well-known one-stage object detector based on anchoring technique. We slightly modified the original anchors configuration by using an anchor box size ranging from $8^2$ to $128^2$ in order to be more suitable for small lesion detection. We applied it to the whole image without any kind of rescale or patching.

We applied a statistical comparison by means of bootstrap method \citep{Bootstrapping_2006} to test the significance of observed performances. Cases were sampled with replacement $1,000$ times, with each bootstrap containing the same number of cases as the original set. At each bootstrapping iteration, FROC curves were recalculated, and differences in figures-of-merit $\Delta AUFC_\gamma$ between \textit{GravityNet} and each of the compared methods were evaluated. 
Finally, the obtained FROC curves were averaged along the TPR axis, and $p$-values were computed as the fraction of $\Delta AUF_\gamma$ populations that were negative or zero. The statistical significance level was chosen as $\alpha = 0.05$. Average FROC curve are shown in Fig.~\ref{fig:FROC_ResultsComparison}.

The statistical comparison results for MCs and MAs detection are shown in Tab.~\ref{tab:ComparisonResults} where significant performances are indicated in bold.
Results of the proposed architecture were statistically significantly better than all the other considered approaches. The highest improvement in terms of $AUFC_\gamma$ is of $+ 50.04 \%$ with RetinaNet for MAs and of $+ 42.25 \%$ with CSNet for MCs. Compared to patch-based methods such as DC, DC-MCNet and ME-MCNet the improvement is respectively $+ 41.00 \%$, $+ 19.52 \%$, $+ 11.90 \%$ for MCs and $+ 15.72 \%$, $+ 10.15 \%$, $+ 5.90 \%$ for MAs.

\afterpage{
    \begin{table*}[t!]
    \caption{Results comparison in terms \% of $AUFC_\gamma$}
    \vspace{0.3cm}
    \centering
    \begin{adjustbox}{scale=0.9}
        \begin{tabular}{c  c  c  c  c  c}
                \hline
                \rule{0pt}{10pt} & \textbf{Method} & \textbf{$AUFC_\gamma$} & \textbf{Compared to} & \textbf{$\Delta AUFC_\gamma$} & \textbf{p-Value} \\ [4pt]
                \hline
                \multicolumn{1}{c}{\multirow{12}{*}{\textbf{MCs detection}}} \hspace{0.5cm} & \multicolumn{1}{l}{RetinaNet} & 66.47 & & & \\ [6pt]
                & \multicolumn{1}{l}{CSNet} & 30.00 & & & \\ [6pt]
                & \multicolumn{1}{l}{DC} & 31.25 & & & \\ [6pt]
                & \multicolumn{1}{l}{DC-MCNet} & 52.73 & & & \\ [6pt]
                & \multicolumn{1}{l}{ME-MCNet} & 60.35 & & & \\ [6pt]
                \cmidrule(lr){2-6}
                & \multicolumn{1}{l}{\textbf{GravityNet}} & \textbf{72.25} & \multicolumn{1}{l}{RetinaNet} & \textbf{+5.78} & $= 0.037$\\ [6pt]
                &            &        & \multicolumn{1}{l}{CSNet} & \textbf{+42.25} & $< 0.001$ \\ [6pt]
                &            &        & \multicolumn{1}{l}{DC} & \textbf{+41} & $< 0.001$ \\ [6pt]
                &            &        & \multicolumn{1}{l}{DC-MCNet} & \textbf{+19.52} & $< 0.001$ \\ [6pt]
                &            &        & \multicolumn{1}{l}{ME-MCNet} & \textbf{+11.9} & $< 0.001$ \\ [6pt]
                \hline
                \multicolumn{1}{c}{\multirow{12}{*}{\textbf{MAs detection}}} \hspace{0.5cm} & \multicolumn{1}{l}{RetinaNet} & 21.48 & & & \\ [6pt]
                & \multicolumn{1}{l}{CSNet} & 40.03 & & & \\ [6pt]
                & \multicolumn{1}{l}{DC} & 55.80 & & & \\ [6pt]
                & \multicolumn{1}{l}{DC-MCNet} & 61.38 & & & \\ [6pt]
                & \multicolumn{1}{l}{ME-MCNet} & 65.63 & & & \\ [6pt]
                \cmidrule(lr){2-6}
                & \multicolumn{1}{l}{\textbf{GravityNet}} & \textbf{71.53} & \multicolumn{1}{l}{RetinaNet} & \textbf{+50.04} & $< 0.001$ \\ [6pt]
                &            &        & \multicolumn{1}{l}{CSNet} & \textbf{+31.49} & $< 0.001$ \\ [6pt]
                &            &        & \multicolumn{1}{l}{DC} & \textbf{+15.72} & $< 0.001$ \\ [6pt]
                &            &        & \multicolumn{1}{l}{DC-MCNet} & \textbf{+10.15} & $< 0.001$ \\ [6pt]
                &            &        & \multicolumn{1}{l}{ME-MCNet} & \textbf{+5.9} & $< 0.001$ \\ [6pt]
                \hline
            \label{tab:ComparisonResults}
        \end{tabular}
    \end{adjustbox}
    \end{table*}
    
    \begin{figure*}[h!]
    \centering
    \begin{subfigure}[t]{0.45\textwidth}
        \includegraphics[width=\linewidth]{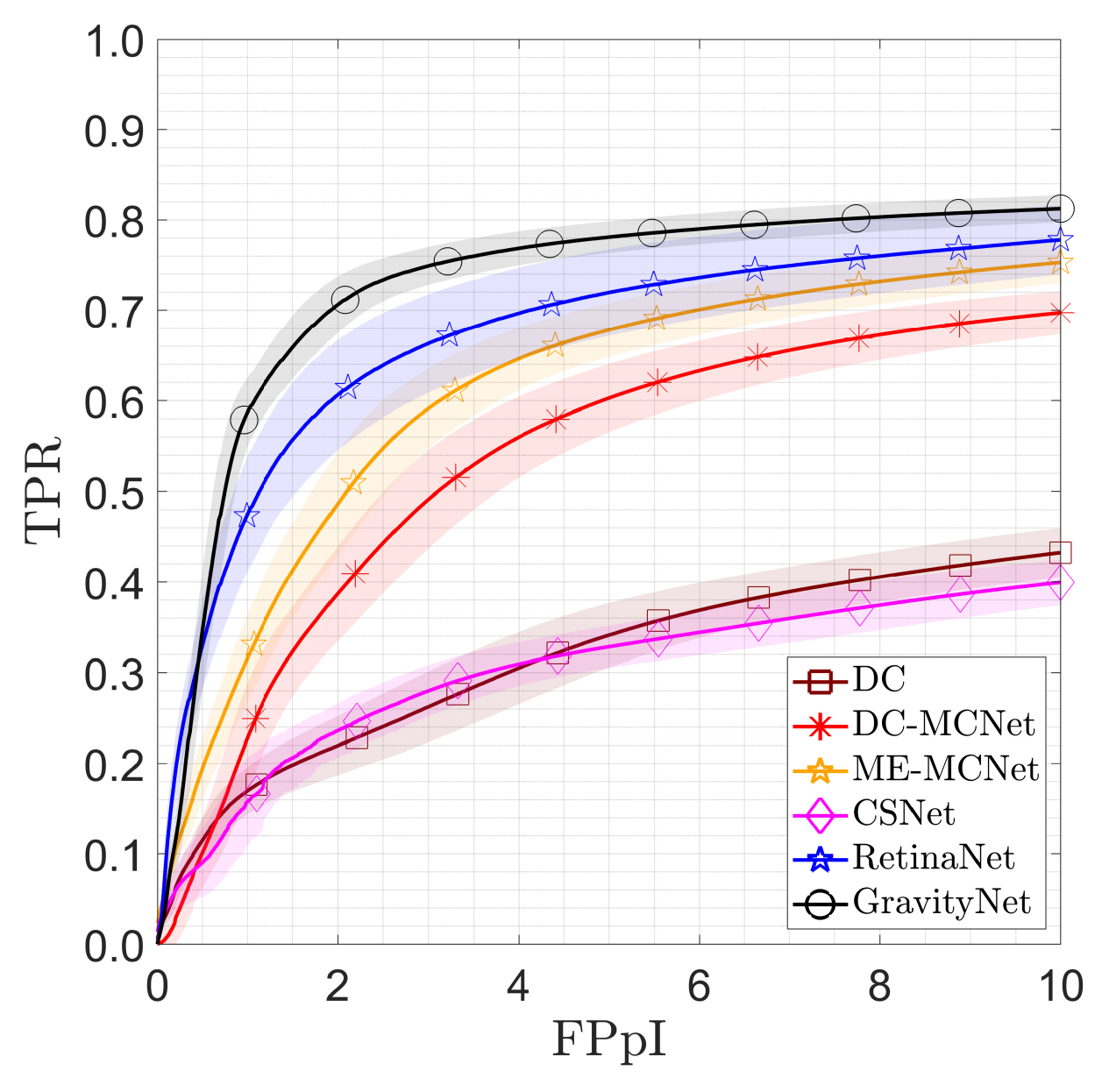}
        \caption{}
        \label{subfig:FROC_ResultsComparison_MCs}
    \end{subfigure}\hspace{0.3cm} 
    \begin{subfigure}[t]{0.45\textwidth}
        \includegraphics[width=\linewidth]{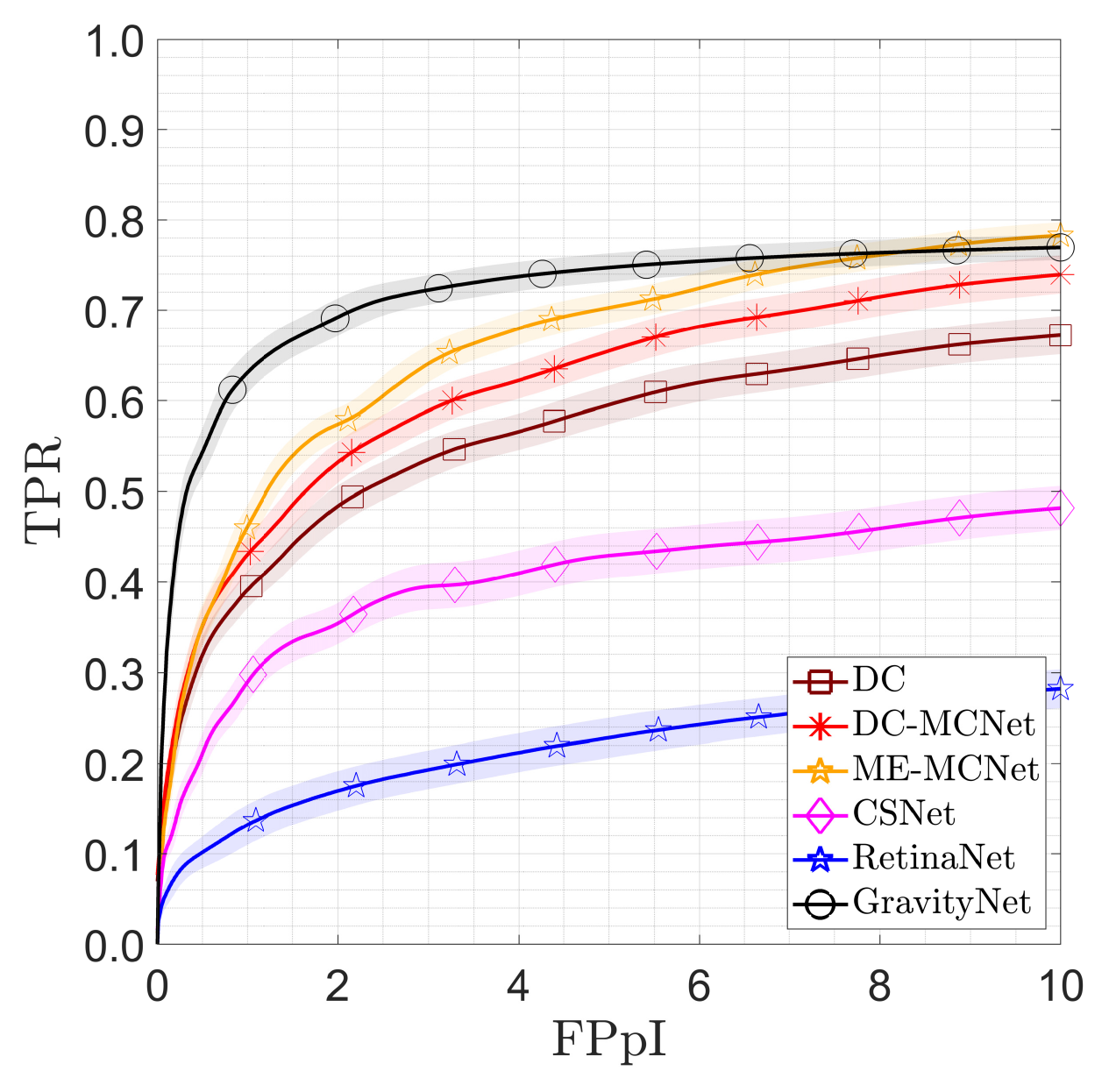}
        \caption{}
        \label{subfig:FROC_ResultsComparison_MAs}
    \end{subfigure}
    \caption{Average FROC curves for INbreast (a) and E-ophtha-MA (b) obtained from $1,000$ bootstrap iterations. Confidence bands (semi-transparent) indicate $95\%$ confidence intervals along the TPR axis.}
    \label{fig:FROC_ResultsComparison}
    \end{figure*}
}

\section{Discussion}\label{sec:discussion}

\subsection{Gravity points configuration}
The gravity points configuration depends directly on the size of the input image and is managed by the \textit{step} parameter. This implies a higher number of gravity points for images with larger dimensions. In the cases studied, mammograms have a larger size than retina images and consequently have a higher $N_{GP}$, so requiring much more computational efforts.

Depending on the chosen configuration, gravity-points will behave differently. We chose to train all the configurations with a $d_h$ equal to the \textit{step} to measure the capacity of gravity-points to move towards ground-truth lesions. A small $d_h$ will have less impact on the movement of gravity-points, compared to a large $d_h$ that let them move more widely, always within the specified distance value. For MCs detection, the best configurations are those with \textit{step} $10$ and thus $d_h$ $10$ because these values are more representative of the size and distribution of MCs in mammographies. On the other hand, for MAs detection, where lesions are usually isolated, configurations with a higher density, such as \textit{step} $5$ and \textit{step} $6$, are needed. Fig.~\ref{fig:output} shows two detection outputs of the best GravityNet models for MCs with \textit{step} $10$ and \textit{ResNet-34} and for MAs with \textit{step} $6$ and \textit{ResNet-50}. We can see the gravitational behaviour towards the centres of the lesions in Fig.~\ref{subfig:output_MCs} for MCs and Fig.~\ref{subfig:output_MAs} for MAs. Hooked gravity-points that are, at inference time, within the radius of the lesion to be detected are shown in light blue and are defined as predictions of possible TP. The NMS, whose output can be seen in the right panel of the same figures, merges all hooked gravity-points in a single detection so as to obtain a single prediction (in blue) for each small lesion (in green).

\afterpage{
    \begin{figure*}[h!]
    \vspace{0.5cm}
    \centering
    \begin{subfigure}[t]{0.45\textwidth}
        \includegraphics[width=\textwidth]{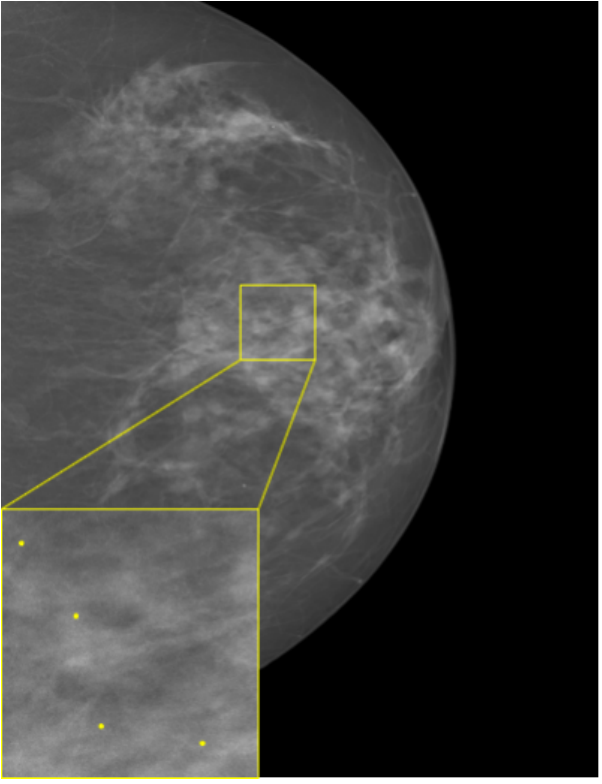}
        \caption{}
        \label{subfig:output_MCs_gt}
    \end{subfigure}\hspace{1.0cm}\vspace{1.0cm} 
    \begin{subfigure}[t]{0.45\textwidth}
        \includegraphics[width=\textwidth]{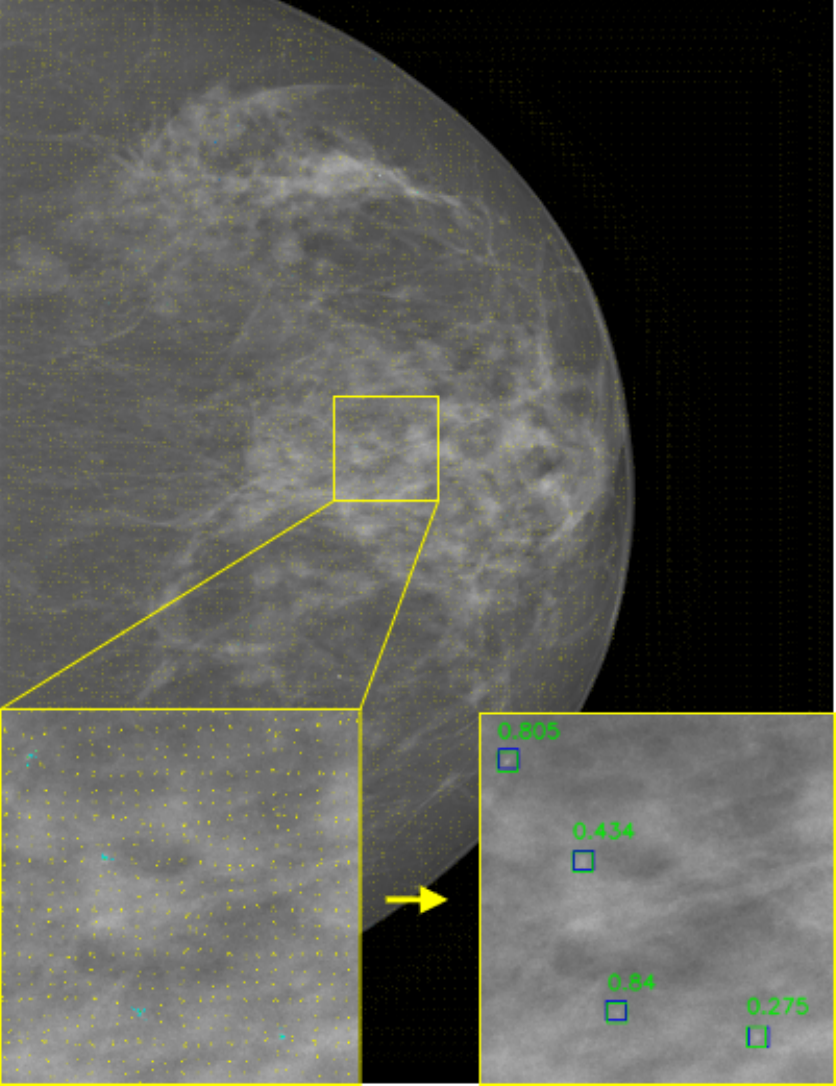}
        \caption{}
        \label{subfig:output_MCs}
    \end{subfigure}
    \begin{subfigure}[t]{0.45\textwidth}
        \includegraphics[width=\textwidth]{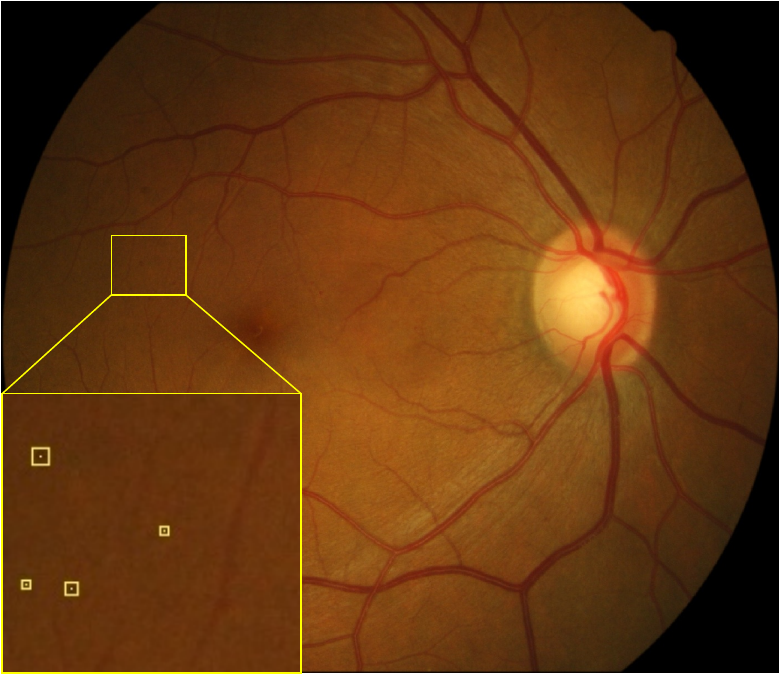}
        \caption{}
        \label{subfig:output_MAs_gt}
    \end{subfigure}\hspace{1.0cm}\vspace{1.0cm} 
    \begin{subfigure}[t]{0.45\textwidth}
        \includegraphics[width=\linewidth]{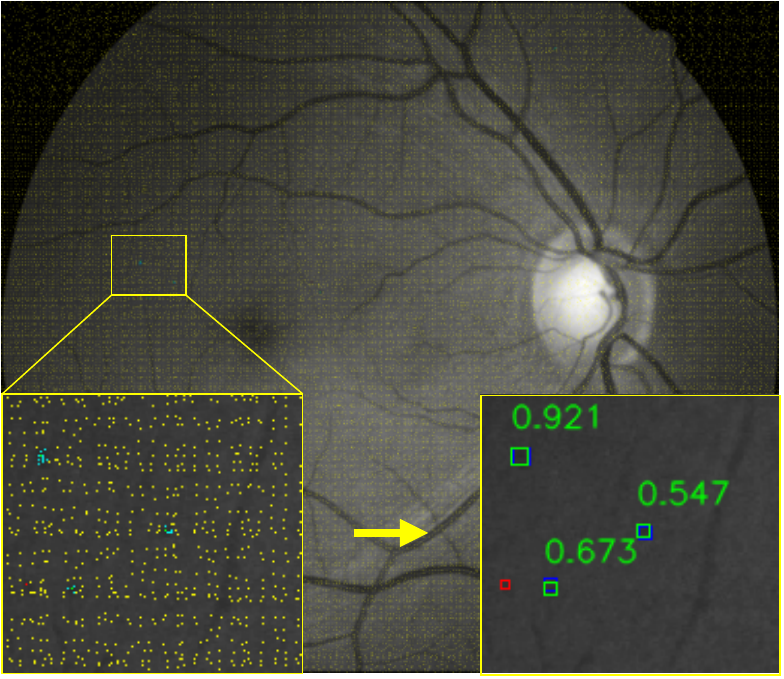}
        \caption{}
        \label{subfig:output_MAs}
    \end{subfigure}
    \caption{Examples of MCs and MAs detections. (a) and (c): ground-truth annotations; (b) and (d): GravityNet outputs}
    \label{fig:output}
    \end{figure*}
\clearpage\clearpage
}

\subsection{Comparison with anchoring methods}
We compared the proposed one-stage detector with a widespread exponent of one-stage object detection methods, i.e. \textit{RetinaNet}, which has also been usefully applied to medical detection problems \citep{Jung_et_al_2018, Lotter_et_al_2021}. Small lesions such as MCs and MAs are often less than 10 pixels in diameter and, in this case, anchoring methods face two main obstacles: (i) the number and size of anchor boxes, and (ii) the pyramidal approach for multi-scale resolution.

Regarding the first issue, we tried to train \textit{RetinaNet} with the original range of anchor boxes size (from $32^2$ to $512^2$ according to the \textit{Feature Pyramid Network} (FPN) level), but due to the small size of the lesions, the train failed; thus, we reduced the size in the range $8^2$ to $128^2$. The proposed anchoring technique is based on pixel-shaped gravity points, which only require an initial configuration setting without specifying a box size. This is advantageous, especially in the case of small lesions with variable sizes, as demonstrated in the MA results. In addition, considering all FPN resolution levels, \textit{RetinaNet} generates a number of anchor boxes more than 10 times the number of gravity points. This is a considerable advantage in computational and temporal terms (see Section~\ref{sec:ComputationalTime}).

As to the second issue, \textit{RetinaNet} adopts a multi-scale architecture. However, this approach proves to be ineffective because positive anchors (those containing a lesion) only belong to the first level of FPN, which corresponds to the highest resolution level. In \textit{GravityNet}, we decided to not use a multi-scale approach given the shape of the lesions to be detected. For the sake of comparison, we tried to use \textit{RetinaNet} without FPN, considering only the outputs of the first level, but this did not improve the performance.

\subsection{Comparison with patch-based methods}
Patch-based methods have the computational disadvantage of assembling all the individual results to obtain the final one, as opposed to end-to-end systems like \textit{GravityNet} that obtain the final result directly.

The class imbalance between lesions and background is another issue that affects small lesion detection. We compared our approach with two existing methods, \textit{DC} and \textit{DC-MCNet}, which are designed to manage this problem. \textit{DC} discards the majority of easily detectable background samples early in the process, while \textit{DC-MCNet} utilizes a CNN on the output of \textit{DC} to enhance detection performance. In this work, we propose the \textit{Gravity Loss}, a variant of \textit{Focal Loss} typically applied in deep learning methods to address class imbalance issues.

Since small lesions do not have a clear appearance and are similar to the surrounding background, \textit{CSNet} and \textit{ME-MCNet} propose two context-sensitive patch-based approaches, where the model is trained with patches of different sizes and then combined. In contrast, our proposal works with the full image without patches and is able to identify small lesions thanks to the new anchoring technique and the regression subnet, which focus more on the distance to the lesion rather than its appearance.

\subsection{Computational and inference time}\label{sec:ComputationalTime}
Computational and inference time are an important aspect in medical imaging systems, improving interactivity and the time taken to formulate a diagnosis.
We evaluated the computational time for all the compared methods by measuring the average \textit{Time per Epoch} (TpE) in training, and the \textit{Time per Image} (TpI) and the Throughput\footnote{Throughput is defined as the maximum number of input instances that the method can process in one second} in test. Tab.~\ref{tab:TimeComparisonResults} shows the results. We can see how patch-based methods are computationally time-consuming, whereas our proposal has a very high Throughput and an average TpI below one second.

\begin{table}[h!]
\caption{Computational times compared in terms of \textit{Time per Epoch} (TpE) in training, and \textit{Time per Image} (TpI) and \textit{Throughput} in test}
\vspace{0.3cm}
\centering
\begin{adjustbox}{scale=0.7}
    \begin{tabular}{c  c  c  c  c}
            \hline
            \rule{0pt}{10pt} & \textbf{Method} & \textbf{TpE (s)} & \textbf{TpI} (s) & \textbf{Throughput} \\ [4pt]
            \hline
            \multicolumn{1}{c}{\multirow{8}{*}{\textbf{MCs detection}}} \hspace{0.5cm} & \multicolumn{1}{l}{RetinaNet} & 1254 & 0.121 & 14.70 \\ [6pt]
            & \multicolumn{1}{l}{CSNet} & 6959 & 822 & $1.2 \times 10^{-3}$ \\ [6pt]
            & \multicolumn{1}{l}{DC} & n.a. & 1.1 & 0.9 \\ [6pt]
            & \multicolumn{1}{l}{DC-MCNet} & 32 & 1.2 & 0.8 \\ [6pt]
            & \multicolumn{1}{l}{ME-MCNet} & 5360 & 386 & $3.8 \times 10^{-3}$ \\ [6pt]
            \cmidrule(lr){2-5}
            & \multicolumn{1}{l}{\textbf{GravityNet}} & \textbf{607} & \textbf{0.061} & \textbf{19.25} \\ [6pt]
            \hline
            \multicolumn{1}{c}{\multirow{8}{*}{\textbf{MAs detection}}} \hspace{0.5cm} & \multicolumn{1}{l}{RetinaNet} & 137 & 0.057 & 34.04 \\ [6pt]
            & \multicolumn{1}{l}{CSNet} & 1260 & 266 & $3.8 \times 10^{-3}$ \\ [6pt]
            & \multicolumn{1}{l}{DC} & n.a. & 1.3 & 0.7 \\ [6pt]
            & \multicolumn{1}{l}{DC-MCNet} & 6 & 1.4 & 0.7 \\ [6pt]
            & \multicolumn{1}{l}{ME-MCNet} & 1564 & 203 & $4.9 \times 10^{-3}$ \\ [6pt]
            \cmidrule(lr){2-5}
            & \multicolumn{1}{l}{\textbf{GravityNet}} & \textbf{184} & \textbf{0.045} & \textbf{37.49} \\ [6pt]
            \hline
        \label{tab:TimeComparisonResults}
    \end{tabular}
\end{adjustbox}
\end{table}

\subsection{Limitations}
Although our method achieves excellent results in the detection of small lesions, there are some limitations to be considered:
\begin{itemize}
    \item[-] Clinical applicability: we require a dataset with individually annotated lesions for the training phase, and this can be difficult to meet in a real clinical scenario. In addition, further post-processing (e.g. benign vs. malignant lesion classification) is needed to build a full CAD system.
    \item [-] Configuration limit: by employing an equispaced grid configuration, the distribution of gravity points becomes uniform, even in areas of the image where there is no tissue. In training this might not be advantageous. Different approaches to generate the initial configuration can be investigated.
    \item[-] Computational requirements: the number of gravity points directly increases with the size of the image. In case of large images, \textit{GravityNet} can require considerable computational resources. A solution can be to limit the number of gravity points by using sparse initial configuration, but this can affect the detection performance of the method.
    \item [-] Memory constraints: the use of a backbone in the proposed model necessitates remarkable resource requirements. As the backbone architecture becomes more complex and deeper, it requires a larger memory allocation, which can be a significant limitation for training the model.
\end{itemize}

\section{Conclusions and future work}\label{sec:conclusions}
In this work, we introduced \textit{GravityNet}, a new one-stage end-to-end detector specifically designed to detect small lesions in medical images. The accurate localization of small lesions, given their appearance and diverse contextual backgrounds, is a challenge in several medical applications. To address this point, our approach employed a novel pixel-based anchor that dynamically moves towards the targeted lesion during detection. Through a comparative evaluation with state-of-the-art anchoring and patch-based methods, our proposed approach demonstrated promising results in effectively detecting small lesions.

Our primary future direction will involve testing \textit{GravityNet} in various detection problems, particularly those where the target object is point-like, such as nuclei localization in whole-slide images \citep{Jiang_Zhou_et_al_2023}. We will also explore the possibility of extending the proposed architecture to address other tasks or image dimensionality involving small lesions, such as segmentation \citep{Li_et_al_2021} or three-dimensional images \citep{Kern_et_al_2021, Toosi_et_al_23}.

\printcredits

\bibliographystyle{cas-model2-names}

\bibliography{refs}

\begin{thebibliography}{68}
\expandafter\ifx\csname natexlab\endcsname\relax\def\natexlab#1{#1}\fi
\providecommand{\url}[1]{\texttt{#1}}
\providecommand{\href}[2]{#2}
\providecommand{\path}[1]{#1}
\providecommand{\DOIprefix}{doi:}
\providecommand{\ArXivprefix}{arXiv:}
\providecommand{\URLprefix}{URL: }
\providecommand{\Pubmedprefix}{pmid:}
\providecommand{\doi}[1]{\href{http://dx.doi.org/#1}{\path{#1}}}
\providecommand{\Pubmed}[1]{\href{pmid:#1}{\path{#1}}}
\providecommand{\bibinfo}[2]{#2}
\ifx\xfnm\relax \def\xfnm[#1]{\unskip,\space#1}\fi
\bibitem[{Bengio et~al.(1994)Bengio, Simard and
  Frasconi}]{Bengio_Simard_Frasconi_1994}
\bibinfo{author}{Bengio, Y.}, \bibinfo{author}{Simard, P.},
  \bibinfo{author}{Frasconi, P.}, \bibinfo{year}{1994}.
\newblock \bibinfo{title}{Learning long-term dependencies with gradient descent
  is difficult}.
\newblock \bibinfo{journal}{IEEE Transactions on Neural Networks}
  \bibinfo{volume}{5}, \bibinfo{pages}{157–166}.
\newblock \DOIprefix\doi{10.1109/72.279181}.
\bibitem[{Bria et~al.(2016)Bria, Marrocco, Karssemeijer, Molinara and
  Tortorella}]{Bria_et_al_2016}
\bibinfo{author}{Bria, A.}, \bibinfo{author}{Marrocco, C.},
  \bibinfo{author}{Karssemeijer, N.}, \bibinfo{author}{Molinara, M.},
  \bibinfo{author}{Tortorella, F.}, \bibinfo{year}{2016}.
\newblock \bibinfo{title}{Deep cascade classifiers to detect clusters of
  microcalcifications}, in: \bibinfo{editor}{Tingberg, A.},
  \bibinfo{editor}{Lång, K.}, \bibinfo{editor}{Timberg, P.} (Eds.),
  \bibinfo{booktitle}{Breast Imaging}, \bibinfo{publisher}{Springer
  International Publishing}. p. \bibinfo{pages}{415–422}.
\newblock \DOIprefix\doi{10.1007/978-3-319-41546-8_52}.
\bibitem[{Bria et~al.(2020)Bria, Marrocco and Tortorella}]{Bria_et_al_2020}
\bibinfo{author}{Bria, A.}, \bibinfo{author}{Marrocco, C.},
  \bibinfo{author}{Tortorella, F.}, \bibinfo{year}{2020}.
\newblock \bibinfo{title}{Addressing class imbalance in deep learning for small
  lesion detection on medical images}.
\newblock \bibinfo{journal}{Computers in Biology and Medicine}
  \bibinfo{volume}{120}, \bibinfo{pages}{103735}.
\newblock \DOIprefix\doi{10.1016/j.compbiomed.2020.103735}.
\bibitem[{Chakraborty(2008)}]{Chakraborty_2008}
\bibinfo{author}{Chakraborty, D.P.}, \bibinfo{year}{2008}.
\newblock \bibinfo{title}{Validation and statistical power comparison of
  methods for analyzing free-response observer performance studies}.
\newblock \bibinfo{journal}{Academic Radiology} \bibinfo{volume}{15},
  \bibinfo{pages}{1554–1566}.
\newblock \DOIprefix\doi{https://doi.org/10.1016/j.acra.2008.07.018}.
\bibitem[{Chen et~al.(2022a)Chen, Duan, Wang, Wang, Li, Qi, Duan and
  Qi}]{Chen_Duan_et_al_2022}
\bibinfo{author}{Chen, S.}, \bibinfo{author}{Duan, J.}, \bibinfo{author}{Wang,
  H.}, \bibinfo{author}{Wang, R.}, \bibinfo{author}{Li, J.},
  \bibinfo{author}{Qi, M.}, \bibinfo{author}{Duan, Y.}, \bibinfo{author}{Qi,
  S.}, \bibinfo{year}{2022}a.
\newblock \bibinfo{title}{Automatic detection of stroke lesion from
  diffusion-weighted imaging via the improved yolov5}.
\newblock \bibinfo{journal}{Computers in Biology and Medicine}
  \bibinfo{volume}{150}, \bibinfo{pages}{106120}.
\newblock \DOIprefix\doi{https://doi.org/10.1016/j.compbiomed.2022.106120}.
\bibitem[{Chen et~al.(2022b)Chen, Wang, Zhang, Fung, Thai, Moore, Mannel, Liu,
  Zheng and Qiu}]{Chen_et_al_2022}
\bibinfo{author}{Chen, X.}, \bibinfo{author}{Wang, X.}, \bibinfo{author}{Zhang,
  K.}, \bibinfo{author}{Fung, K.M.}, \bibinfo{author}{Thai, T.C.},
  \bibinfo{author}{Moore, K.}, \bibinfo{author}{Mannel, R.S.},
  \bibinfo{author}{Liu, H.}, \bibinfo{author}{Zheng, B.}, \bibinfo{author}{Qiu,
  Y.}, \bibinfo{year}{2022}b.
\newblock \bibinfo{title}{Recent advances and clinical applications of deep
  learning in medical image analysis}.
\newblock \bibinfo{journal}{Medical Image Analysis} \bibinfo{volume}{79},
  \bibinfo{pages}{102444}.
\newblock \DOIprefix\doi{10.1016/j.media.2022.102444}.
\bibitem[{Ciga et~al.(2021)Ciga, Xu, Nofech-Mozes, Noy, Lu and
  Martel}]{Ciga_et_al_2021}
\bibinfo{author}{Ciga, O.}, \bibinfo{author}{Xu, T.},
  \bibinfo{author}{Nofech-Mozes, S.}, \bibinfo{author}{Noy, S.},
  \bibinfo{author}{Lu, F.I.}, \bibinfo{author}{Martel, A.L.},
  \bibinfo{year}{2021}.
\newblock \bibinfo{title}{Overcoming the limitations of patch-based learning to
  detect cancer in whole slide images}.
\newblock \bibinfo{journal}{Scientific Reports} \bibinfo{volume}{11},
  \bibinfo{pages}{8894}.
\newblock \DOIprefix\doi{10.1038/s41598-021-88494-z}.
\bibitem[{Civilibal et~al.(2023)Civilibal, Cevik and
  Bozkurt}]{Civilibal_et_al_2023}
\bibinfo{author}{Civilibal, S.}, \bibinfo{author}{Cevik, K.K.},
  \bibinfo{author}{Bozkurt, A.}, \bibinfo{year}{2023}.
\newblock \bibinfo{title}{A deep learning approach for automatic detection,
  segmentation and classification of breast lesions from thermal images}.
\newblock \bibinfo{journal}{Expert Systems with Applications}
  \bibinfo{volume}{212}, \bibinfo{pages}{118774}.
\newblock \DOIprefix\doi{10.1016/j.eswa.2022.118774}.
\bibitem[{Dashtbozorg et~al.(2018)Dashtbozorg, Zhang, Huang and ter
  Haar~Romeny}]{Dashtbozorg_et_al_2018}
\bibinfo{author}{Dashtbozorg, B.}, \bibinfo{author}{Zhang, J.},
  \bibinfo{author}{Huang, F.}, \bibinfo{author}{ter Haar~Romeny, B.M.},
  \bibinfo{year}{2018}.
\newblock \bibinfo{title}{Retinal microaneurysms detection using local
  convergence index features}.
\newblock \bibinfo{journal}{IEEE Transactions on Image Processing}
  \bibinfo{volume}{27}, \bibinfo{pages}{3300–3315}.
\newblock \DOIprefix\doi{10.1109/TIP.2018.2815345}.
\bibitem[{Dass and Kumar(2022)}]{Dass_et_al_2022}
\bibinfo{author}{Dass, J.M.A.}, \bibinfo{author}{Kumar, S.M.},
  \bibinfo{year}{2022}.
\newblock \bibinfo{title}{A novel approach for small object detection in
  medical images through deep ensemble convolution neural network}.
\newblock \bibinfo{journal}{International Journal of Advanced Computer Science
  and Applications (IJACSA)} \bibinfo{volume}{13}.
\newblock \DOIprefix\doi{10.14569/IJACSA.2022.0130380}.
\bibitem[{Decencière et~al.(2013)Decencière, Cazuguel, Zhang, Thibault,
  Klein, Meyer, Marcotegui, Quellec, Lamard, Danno, Elie, Massin, Viktor,
  Erginay, Laÿ and Chabouis}]{Eophtha_2013}
\bibinfo{author}{Decencière, E.}, \bibinfo{author}{Cazuguel, G.},
  \bibinfo{author}{Zhang, X.}, \bibinfo{author}{Thibault, G.},
  \bibinfo{author}{Klein, J.C.}, \bibinfo{author}{Meyer, F.},
  \bibinfo{author}{Marcotegui, B.}, \bibinfo{author}{Quellec, G.},
  \bibinfo{author}{Lamard, M.}, \bibinfo{author}{Danno, R.},
  \bibinfo{author}{Elie, D.}, \bibinfo{author}{Massin, P.},
  \bibinfo{author}{Viktor, Z.}, \bibinfo{author}{Erginay, A.},
  \bibinfo{author}{Laÿ, B.}, \bibinfo{author}{Chabouis, A.},
  \bibinfo{year}{2013}.
\newblock \bibinfo{title}{Teleophta: Machine learning and image processing
  methods for teleophthalmology}.
\newblock \bibinfo{journal}{IRBM} \bibinfo{volume}{34},
  \bibinfo{pages}{196--203}.
\newblock \DOIprefix\doi{10.1016/j.irbm.2013.01.010}.
\bibitem[{Ding et~al.(2017)Ding, Li, Hu and Wang}]{Ding_et_all_2017}
\bibinfo{author}{Ding, J.}, \bibinfo{author}{Li, A.}, \bibinfo{author}{Hu, Z.},
  \bibinfo{author}{Wang, L.}, \bibinfo{year}{2017}.
\newblock \bibinfo{title}{Accurate pulmonary nodule detection in computed
  tomography images using deep convolutional neural networks}, in:
  \bibinfo{editor}{Descoteaux, M.}, \bibinfo{editor}{Maier-Hein, L.},
  \bibinfo{editor}{Franz, A.}, \bibinfo{editor}{Jannin, P.},
  \bibinfo{editor}{Collins, D.L.}, \bibinfo{editor}{Duchesne, S.} (Eds.),
  \bibinfo{booktitle}{Medical Image Computing and Computer Assisted
  Intervention MICCAI 2017}, \bibinfo{publisher}{Springer International
  Publishing}, \bibinfo{address}{Cham}. p. \bibinfo{pages}{559–567}.
\newblock \DOIprefix\doi{10.1007/978-3-319-66179-7_64}.
\bibitem[{Eadie et~al.(2012)Eadie, Taylor and Gibson}]{Eadie_et_al_2012}
\bibinfo{author}{Eadie, L.H.}, \bibinfo{author}{Taylor, P.},
  \bibinfo{author}{Gibson, A.P.}, \bibinfo{year}{2012}.
\newblock \bibinfo{title}{A systematic review of computer-assisted diagnosis in
  diagnostic cancer imaging}.
\newblock \bibinfo{journal}{European Journal of Radiology}
  \bibinfo{volume}{81}, \bibinfo{pages}{e70–e76}.
\newblock \DOIprefix\doi{10.1016/j.ejrad.2011.01.098}.
\bibitem[{Ezra et~al.(2013)Ezra, Keinan, Mandel, Boulton and
  Nahmias}]{Ezra_et_al_2013}
\bibinfo{author}{Ezra, E.}, \bibinfo{author}{Keinan, E.},
  \bibinfo{author}{Mandel, Y.}, \bibinfo{author}{Boulton, M.E.},
  \bibinfo{author}{Nahmias, Y.}, \bibinfo{year}{2013}.
\newblock \bibinfo{title}{Non-dimensional analysis of retinal microaneurysms:
  critical threshold for treatment}.
\newblock \bibinfo{journal}{Integrative biology: quantitative biosciences from
  nano to macro} \bibinfo{volume}{5}, \bibinfo{pages}{474–480}.
\newblock \DOIprefix\doi{10.1039/c3ib20259c}.
\bibitem[{Girshick(2015)}]{Faster_RCNN_2015}
\bibinfo{author}{Girshick, R.}, \bibinfo{year}{2015}.
\newblock \bibinfo{title}{Fast r-cnn}, in: \bibinfo{booktitle}{2015 IEEE
  International Conference on Computer Vision (ICCV)}, pp.
  \bibinfo{pages}{1440--1448}.
\newblock \DOIprefix\doi{10.1109/ICCV.2015.169}.
\bibitem[{Glorot and Bengio(2010)}]{Xavier_2010}
\bibinfo{author}{Glorot, X.}, \bibinfo{author}{Bengio, Y.},
  \bibinfo{year}{2010}.
\newblock \bibinfo{title}{Understanding the difficulty of training deep
  feedforward neural networks}, in: \bibinfo{booktitle}{Proceedings of the
  Thirteenth International Conference on Artificial Intelligence and
  Statistics}, \bibinfo{publisher}{JMLR Workshop and Conference Proceedings}.
  p. \bibinfo{pages}{249–256}.
\bibitem[{Gu et~al.(2022)Gu, Wu, Wu, Wang, Yang, Chen, Wang and
  Chen}]{Gu_et_all_2022}
\bibinfo{author}{Gu, F.}, \bibinfo{author}{Wu, X.}, \bibinfo{author}{Wu, W.},
  \bibinfo{author}{Wang, Z.}, \bibinfo{author}{Yang, X.},
  \bibinfo{author}{Chen, Z.}, \bibinfo{author}{Wang, Z.},
  \bibinfo{author}{Chen, G.}, \bibinfo{year}{2022}.
\newblock \bibinfo{title}{Performance of deep learning in the detection of
  intracranial aneurysm: A systematic review and meta-analysis}.
\newblock \bibinfo{journal}{European Journal of Radiology}
  \bibinfo{volume}{155}, \bibinfo{pages}{110457}.
\newblock \DOIprefix\doi{10.1016/j.ejrad.2022.110457}.
\bibitem[{Han et~al.(2022)Han, Liu and Chen}]{Han_et_al_2022}
\bibinfo{author}{Han, R.}, \bibinfo{author}{Liu, X.}, \bibinfo{author}{Chen,
  T.}, \bibinfo{year}{2022}.
\newblock \bibinfo{title}{Yolo-sg: Salience-guided detection of small objects
  in medical images}, in: \bibinfo{booktitle}{2022 IEEE International
  Conference on Image Processing (ICIP)}, pp. \bibinfo{pages}{4218--4222}.
\newblock \DOIprefix\doi{10.1109/ICIP46576.2022.9898077}.
\bibitem[{He et~al.(2017)He, Gkioxari, Dollár and Girshick}]{MaskRCNN_2017}
\bibinfo{author}{He, K.}, \bibinfo{author}{Gkioxari, G.},
  \bibinfo{author}{Dollár, P.}, \bibinfo{author}{Girshick, R.},
  \bibinfo{year}{2017}.
\newblock \bibinfo{title}{Mask r-cnn}, in: \bibinfo{booktitle}{2017 IEEE
  International Conference on Computer Vision (ICCV)}, pp.
  \bibinfo{pages}{2980--2988}.
\newblock \DOIprefix\doi{10.1109/ICCV.2017.322}.
\bibitem[{He et~al.(2016)He, Zhang, Ren and Sun}]{ResNet_2015}
\bibinfo{author}{He, K.}, \bibinfo{author}{Zhang, X.}, \bibinfo{author}{Ren,
  S.}, \bibinfo{author}{Sun, J.}, \bibinfo{year}{2016}.
\newblock \bibinfo{title}{Deep residual learning for image recognition}, in:
  \bibinfo{booktitle}{2016 IEEE Conference on Computer Vision and Pattern
  Recognition (CVPR)}, pp. \bibinfo{pages}{770--778}.
\newblock \DOIprefix\doi{10.1109/CVPR.2016.90}.
\bibitem[{Jiang et~al.(2023a)Jiang, Diao, Shi, Zhou, Wang, Hu, Zhu, Luo, Tong
  and Yao}]{Jiang_et_al_2023}
\bibinfo{author}{Jiang, H.}, \bibinfo{author}{Diao, Z.}, \bibinfo{author}{Shi,
  T.}, \bibinfo{author}{Zhou, Y.}, \bibinfo{author}{Wang, F.},
  \bibinfo{author}{Hu, W.}, \bibinfo{author}{Zhu, X.}, \bibinfo{author}{Luo,
  S.}, \bibinfo{author}{Tong, G.}, \bibinfo{author}{Yao, Y.D.},
  \bibinfo{year}{2023}a.
\newblock \bibinfo{title}{A review of deep learning-based multiple-lesion
  recognition from medical images: classification, detection and segmentation}.
\newblock \bibinfo{journal}{Computers in Biology and Medicine}
  \bibinfo{volume}{157}, \bibinfo{pages}{106726}.
\newblock \DOIprefix\doi{10.1016/j.compbiomed.2023.106726}.
\bibitem[{Jiang et~al.(2023b)Jiang, Zhou, Lin, Chan, Liu and
  Chen}]{Jiang_Zhou_et_al_2023}
\bibinfo{author}{Jiang, H.}, \bibinfo{author}{Zhou, Y.}, \bibinfo{author}{Lin,
  Y.}, \bibinfo{author}{Chan, R.C.K.}, \bibinfo{author}{Liu, J.},
  \bibinfo{author}{Chen, H.}, \bibinfo{year}{2023}b.
\newblock \bibinfo{title}{Deep learning for computational cytology: A survey}.
\newblock \bibinfo{journal}{Medical Image Analysis} \bibinfo{volume}{84},
  \bibinfo{pages}{102691}.
\newblock \DOIprefix\doi{10.1016/j.media.2022.102691}.
\bibitem[{Jiao et~al.(2019)Jiao, Zhang, Liu, Yang, Li, Feng and
  Qu}]{Jiao_et_al_2019}
\bibinfo{author}{Jiao, L.}, \bibinfo{author}{Zhang, F.}, \bibinfo{author}{Liu,
  F.}, \bibinfo{author}{Yang, S.}, \bibinfo{author}{Li, L.},
  \bibinfo{author}{Feng, Z.}, \bibinfo{author}{Qu, R.}, \bibinfo{year}{2019}.
\newblock \bibinfo{title}{A survey of deep learning-based object detection}.
\newblock \bibinfo{journal}{IEEE Access} \bibinfo{volume}{7},
  \bibinfo{pages}{128837–128868}.
\newblock \DOIprefix\doi{10.1109/ACCESS.2019.2939201}.
\bibitem[{Jung et~al.(2018)Jung, Kim, Lee, Yoo, Lee, Ham, Woo and
  Kang}]{Jung_et_al_2018}
\bibinfo{author}{Jung, H.}, \bibinfo{author}{Kim, B.}, \bibinfo{author}{Lee,
  I.}, \bibinfo{author}{Yoo, M.}, \bibinfo{author}{Lee, J.},
  \bibinfo{author}{Ham, S.}, \bibinfo{author}{Woo, O.}, \bibinfo{author}{Kang,
  J.}, \bibinfo{year}{2018}.
\newblock \bibinfo{title}{Detection of masses in mammograms using a one-stage
  object detector based on a deep convolutional neural network}.
\newblock \bibinfo{journal}{PLOS ONE} \bibinfo{volume}{13},
  \bibinfo{pages}{e0203355}.
\newblock \DOIprefix\doi{10.1371/journal.pone.0203355}.
\bibitem[{Karimi and Ward(2016)}]{Karimi_Ward_2016}
\bibinfo{author}{Karimi, D.}, \bibinfo{author}{Ward, R.K.},
  \bibinfo{year}{2016}.
\newblock \bibinfo{title}{Patch-based models and algorithms for image
  processing: a review of the basic principles and methods, and their
  application in computed tomography}.
\newblock \bibinfo{journal}{International Journal of Computer Assisted
  Radiology and Surgery} \bibinfo{volume}{11}, \bibinfo{pages}{1765–1777}.
\newblock \DOIprefix\doi{10.1007/s11548-016-1434-z}.
\bibitem[{Kaur and Singh(2022)}]{Kaur_Singh_2022}
\bibinfo{author}{Kaur, R.}, \bibinfo{author}{Singh, S.}, \bibinfo{year}{2022}.
\newblock \bibinfo{title}{A comprehensive review of object detection with deep
  learning}.
\newblock \bibinfo{journal}{Digital Signal Processing} \bibinfo{volume}{132},
  \bibinfo{pages}{103812}.
\newblock \DOIprefix\doi{10.1016/j.dsp.2022.103812}.
\bibitem[{Kawahara and Hamarneh(2016)}]{Kawahara_et_al_2016}
\bibinfo{author}{Kawahara, J.}, \bibinfo{author}{Hamarneh, G.},
  \bibinfo{year}{2016}.
\newblock \bibinfo{title}{Multi-resolution-tract cnn with hybrid pretrained and
  skin-lesion trained layers}, in: \bibinfo{booktitle}{Machine Learning in
  Medical Imaging}, \bibinfo{publisher}{Springer International Publishing}. p.
  \bibinfo{pages}{164–171}.
\newblock \DOIprefix\doi{10.1007/978-3-319-47157-0_20}.
\bibitem[{Kern and Mastmeyer(2021)}]{Kern_et_al_2021}
\bibinfo{author}{Kern, D.}, \bibinfo{author}{Mastmeyer, A.},
  \bibinfo{year}{2021}.
\newblock \bibinfo{title}{3d bounding box detection in volumetric medical image
  data: A systematic literature review}, in: \bibinfo{booktitle}{2021 IEEE 8th
  International Conference on Industrial Engineering and Applications (ICIEA)},
  pp. \bibinfo{pages}{509--516}.
\newblock \DOIprefix\doi{10.1109/ICIEA52957.2021.9436733}.
\bibitem[{Kim et~al.(2020)Kim, Lee, Lee, Jang, Seo and Kim}]{Kim_et_al_2020}
\bibinfo{author}{Kim, Y.G.}, \bibinfo{author}{Lee, S.M.}, \bibinfo{author}{Lee,
  K.H.}, \bibinfo{author}{Jang, R.}, \bibinfo{author}{Seo, J.B.},
  \bibinfo{author}{Kim, N.}, \bibinfo{year}{2020}.
\newblock \bibinfo{title}{Optimal matrix size of chest radiographs for
  computer-aided detection on lung nodule or mass with deep learning}.
\newblock \bibinfo{journal}{European Radiology} \bibinfo{volume}{30},
  \bibinfo{pages}{4943–4951}.
\newblock \DOIprefix\doi{10.1007/s00330-020-06892-9}.
\bibitem[{Kingma and Ba(2017)}]{Adam_2017}
\bibinfo{author}{Kingma, D.P.}, \bibinfo{author}{Ba, J.}, \bibinfo{year}{2017}.
\newblock \bibinfo{title}{Adam: A method for stochastic optimization}.
\newblock \bibinfo{journal}{International Conference on Learning
  Representations (ICLR)} .
\bibitem[{Kisantal et~al.(2019)Kisantal, Wojna, Murawski, Naruniec and
  Cho}]{Kisantal_et_al_2019}
\bibinfo{author}{Kisantal, M.}, \bibinfo{author}{Wojna, Z.},
  \bibinfo{author}{Murawski, J.}, \bibinfo{author}{Naruniec, J.},
  \bibinfo{author}{Cho, K.}, \bibinfo{year}{2019}.
\newblock \bibinfo{title}{Augmentation for small object detection}, in:
  \bibinfo{booktitle}{9th International Conference on Advances in Computing and
  Information Technology (ACITY 2019)}, \bibinfo{publisher}{Aircc Publishing
  Corporation}. p. \bibinfo{pages}{119–133}.
\newblock \DOIprefix\doi{10.5121/csit.2019.91713}.
\bibitem[{LeCun et~al.(2015)LeCun, Bengio and Hinton}]{LeCun_et_al_2015}
\bibinfo{author}{LeCun, Y.}, \bibinfo{author}{Bengio, Y.},
  \bibinfo{author}{Hinton, G.}, \bibinfo{year}{2015}.
\newblock \bibinfo{title}{Deep learning}.
\newblock \bibinfo{journal}{Nature} \bibinfo{volume}{521},
  \bibinfo{pages}{436–444}.
\newblock \DOIprefix\doi{10.1038/nature14539}.
\bibitem[{LeCun et~al.(1998)LeCun, Bottou, Bengio and
  Haffner}]{LeCun_et_al_1998}
\bibinfo{author}{LeCun, Y.}, \bibinfo{author}{Bottou, L.},
  \bibinfo{author}{Bengio, Y.}, \bibinfo{author}{Haffner, P.},
  \bibinfo{year}{1998}.
\newblock \bibinfo{title}{Gradient-based learning applied to document
  recognition}.
\newblock \bibinfo{journal}{Proceedings of the IEEE} \bibinfo{volume}{86},
  \bibinfo{pages}{2278–2324}.
\newblock \DOIprefix\doi{10.1109/5.726791}.
\bibitem[{Lee et~al.(2015)Lee, Wong and Sabanayagam}]{Lee_et_al_2015}
\bibinfo{author}{Lee, R.}, \bibinfo{author}{Wong, T.Y.},
  \bibinfo{author}{Sabanayagam, C.}, \bibinfo{year}{2015}.
\newblock \bibinfo{title}{Epidemiology of diabetic retinopathy, diabetic
  macular edema and related vision loss}.
\newblock \bibinfo{journal}{Eye and Vision} \bibinfo{volume}{2},
  \bibinfo{pages}{17}.
\newblock \DOIprefix\doi{10.1186/s40662-015-0026-2}.
\bibitem[{Li et~al.(2021)Li, Wei, Liu, Atchaneeyasakul, Zhou, Pan, Kumar,
  Zhang, Pu, Liebeskind and Scalzo}]{Li_et_al_2021}
\bibinfo{author}{Li, L.}, \bibinfo{author}{Wei, M.}, \bibinfo{author}{Liu, B.},
  \bibinfo{author}{Atchaneeyasakul, K.}, \bibinfo{author}{Zhou, F.},
  \bibinfo{author}{Pan, Z.}, \bibinfo{author}{Kumar, S.A.},
  \bibinfo{author}{Zhang, J.Y.}, \bibinfo{author}{Pu, Y.},
  \bibinfo{author}{Liebeskind, D.S.}, \bibinfo{author}{Scalzo, F.},
  \bibinfo{year}{2021}.
\newblock \bibinfo{title}{Deep learning for hemorrhagic lesion detection and
  segmentation on brain ct images}.
\newblock \bibinfo{journal}{IEEE Journal of Biomedical and Health Informatics}
  \bibinfo{volume}{25}, \bibinfo{pages}{1646–1659}.
\newblock \DOIprefix\doi{10.1109/JBHI.2020.3028243}.
\bibitem[{Lin et~al.(2017)Lin, Goyal, Girshick, He and
  Dollár}]{RetinaNet_2017}
\bibinfo{author}{Lin, T.Y.}, \bibinfo{author}{Goyal, P.},
  \bibinfo{author}{Girshick, R.}, \bibinfo{author}{He, K.},
  \bibinfo{author}{Dollár, P.}, \bibinfo{year}{2017}.
\newblock \bibinfo{title}{Focal loss for dense object detection}.
\newblock \bibinfo{journal}{IEEE Transactions on Pattern Analysis and Machine
  Intelligence} \bibinfo{volume}{42}, \bibinfo{pages}{318--327}.
\newblock \DOIprefix\doi{10.1109/TPAMI.2018.2858826}.
\bibitem[{Litjens et~al.(2017)Litjens, Kooi, Bejnordi, Setio, Ciompi,
  Ghafoorian, van~der Laak, van Ginneken and Sánchez}]{Litjens_et_al_2017}
\bibinfo{author}{Litjens, G.}, \bibinfo{author}{Kooi, T.},
  \bibinfo{author}{Bejnordi, B.E.}, \bibinfo{author}{Setio, A.A.A.},
  \bibinfo{author}{Ciompi, F.}, \bibinfo{author}{Ghafoorian, M.},
  \bibinfo{author}{van~der Laak, J.A.W.M.}, \bibinfo{author}{van Ginneken, B.},
  \bibinfo{author}{Sánchez, C.I.}, \bibinfo{year}{2017}.
\newblock \bibinfo{title}{A survey on deep learning in medical image analysis}.
\newblock \bibinfo{journal}{Medical Image Analysis} \bibinfo{volume}{42},
  \bibinfo{pages}{60–88}.
\newblock \DOIprefix\doi{10.1016/j.media.2017.07.005}.
\bibitem[{Liu and Deng(2015)}]{VGG16_2015}
\bibinfo{author}{Liu, S.}, \bibinfo{author}{Deng, W.}, \bibinfo{year}{2015}.
\newblock \bibinfo{title}{Very deep convolutional neural network based image
  classification using small training sample size}, in:
  \bibinfo{booktitle}{2015 3rd IAPR Asian Conference on Pattern Recognition
  (ACPR)}, pp. \bibinfo{pages}{730--734}.
\newblock \DOIprefix\doi{10.1109/ACPR.2015.7486599}.
\bibitem[{Liu et~al.(2016)Liu, Anguelov, Erhan, Szegedy, Reed, Fu and
  Berg}]{SSD_2016}
\bibinfo{author}{Liu, W.}, \bibinfo{author}{Anguelov, D.},
  \bibinfo{author}{Erhan, D.}, \bibinfo{author}{Szegedy, C.},
  \bibinfo{author}{Reed, S.}, \bibinfo{author}{Fu, C.Y.},
  \bibinfo{author}{Berg, A.C.}, \bibinfo{year}{2016}.
\newblock \bibinfo{title}{Ssd: Single shot multibox detector}, in:
  \bibinfo{editor}{Leibe, B.}, \bibinfo{editor}{Matas, J.},
  \bibinfo{editor}{Sebe, N.}, \bibinfo{editor}{Welling, M.} (Eds.),
  \bibinfo{booktitle}{Computer Vision – ECCV 2016},
  \bibinfo{publisher}{Springer International Publishing}. p.
  \bibinfo{pages}{21–37}.
\newblock \DOIprefix\doi{0.1007/978-3-319-46448-0_2}.
\bibitem[{Lo et~al.(1995)Lo, Lou, Lin, Freedman, Chien and Mun}]{Lo_et_al_1995}
\bibinfo{author}{Lo, S.C.}, \bibinfo{author}{Lou, S.L.}, \bibinfo{author}{Lin,
  J.S.}, \bibinfo{author}{Freedman, M.}, \bibinfo{author}{Chien, M.},
  \bibinfo{author}{Mun, S.}, \bibinfo{year}{1995}.
\newblock \bibinfo{title}{Artificial convolution neural network techniques and
  applications for lung nodule detection}.
\newblock \bibinfo{journal}{IEEE Transactions on Medical Imaging}
  \bibinfo{volume}{14}, \bibinfo{pages}{711--718}.
\newblock \DOIprefix\doi{10.1109/42.476112}.
\bibitem[{Logullo et~al.(2022)Logullo, Prigenzi, Nimir, Franco and
  Campos}]{Logullo_et_al_2022}
\bibinfo{author}{Logullo, A.F.}, \bibinfo{author}{Prigenzi, K.C.K.},
  \bibinfo{author}{Nimir, C.C.B.A.}, \bibinfo{author}{Franco, A.F.V.},
  \bibinfo{author}{Campos, M.S.D.A.}, \bibinfo{year}{2022}.
\newblock \bibinfo{title}{Breast microcalcifications: Past, present and future
  (review)}.
\newblock \bibinfo{journal}{Molecular and Clinical Oncology}
  \bibinfo{volume}{16}.
\newblock \DOIprefix\doi{10.3892/mco.2022.2514}.
\bibitem[{Lotter et~al.(2021)Lotter, Diab, Haslam, Kim, Grisot, Wu, Wu, Onieva,
  Boyer, Boxerman, Wang, Bandler, Vijayaraghavan and
  Gregory~Sorensen}]{Lotter_et_al_2021}
\bibinfo{author}{Lotter, W.}, \bibinfo{author}{Diab, A.R.},
  \bibinfo{author}{Haslam, B.}, \bibinfo{author}{Kim, J.G.},
  \bibinfo{author}{Grisot, G.}, \bibinfo{author}{Wu, E.}, \bibinfo{author}{Wu,
  K.}, \bibinfo{author}{Onieva, J.O.}, \bibinfo{author}{Boyer, Y.},
  \bibinfo{author}{Boxerman, J.L.}, \bibinfo{author}{Wang, M.},
  \bibinfo{author}{Bandler, M.}, \bibinfo{author}{Vijayaraghavan, G.R.},
  \bibinfo{author}{Gregory~Sorensen, A.}, \bibinfo{year}{2021}.
\newblock \bibinfo{title}{Robust breast cancer detection in mammography and
  digital breast tomosynthesis using an annotation-efficient deep learning
  approach}.
\newblock \bibinfo{journal}{Nature Medicine} \bibinfo{volume}{27},
  \bibinfo{pages}{244–249}.
\newblock \DOIprefix\doi{10.1038/s41591-020-01174-9}.
\bibitem[{Loud and Murphy(2017)}]{Loud_Murphy_2017}
\bibinfo{author}{Loud, J.}, \bibinfo{author}{Murphy, J.}, \bibinfo{year}{2017}.
\newblock \bibinfo{title}{Cancer screening and early detection in the 21st
  century}.
\newblock \bibinfo{journal}{Seminars in oncology nursing} \bibinfo{volume}{33},
  \bibinfo{pages}{121–128}.
\newblock \DOIprefix\doi{10.1016/j.soncn.2017.02.002}.
\bibitem[{Loverdos et~al.(2019)Loverdos, Fotiadis, Kontogianni, Iliopoulou and
  Gaga}]{Loverdos_etl_al_2019}
\bibinfo{author}{Loverdos, K.}, \bibinfo{author}{Fotiadis, A.},
  \bibinfo{author}{Kontogianni, C.}, \bibinfo{author}{Iliopoulou, M.},
  \bibinfo{author}{Gaga, M.}, \bibinfo{year}{2019}.
\newblock \bibinfo{title}{Lung nodules: A comprehensive review on current
  approach and management}.
\newblock \bibinfo{journal}{Annals of Thoracic Medicine} \bibinfo{volume}{14},
  \bibinfo{pages}{226–238}.
\newblock \DOIprefix\doi{10.4103/atm.ATM_110_19}.
\bibitem[{Moreira et~al.(2012)Moreira, Amaral, Domingues, Cardoso, Cardoso and
  Cardoso}]{INbreast_2012}
\bibinfo{author}{Moreira, I.C.}, \bibinfo{author}{Amaral, I.},
  \bibinfo{author}{Domingues, I.}, \bibinfo{author}{Cardoso, A.},
  \bibinfo{author}{Cardoso, M.J.}, \bibinfo{author}{Cardoso, J.S.},
  \bibinfo{year}{2012}.
\newblock \bibinfo{title}{Inbreast}.
\newblock \bibinfo{journal}{Academic Radiology} \bibinfo{volume}{19},
  \bibinfo{pages}{236–248}.
\newblock \DOIprefix\doi{10.1016/j.acra.2011.09.014}.
\bibitem[{Morgan et~al.(2005)Morgan, Cooke and Mccarthy}]{Morgan_et_al_2005}
\bibinfo{author}{Morgan, M.}, \bibinfo{author}{Cooke, M.},
  \bibinfo{author}{Mccarthy, G.}, \bibinfo{year}{2005}.
\newblock \bibinfo{title}{Microcalcifications associated with breast cancer: An
  epiphenomenon or biologically significant feature of selected tumors?}
\newblock \bibinfo{journal}{Journal of mammary gland biology and neoplasia}
  \bibinfo{volume}{10}, \bibinfo{pages}{181–7}.
\newblock \DOIprefix\doi{10.1007/s10911-005-5400-6}.
\bibitem[{Park et~al.(2019)Park, Lee, Kim, Choe, Cho, Do and
  Seo}]{Park_et_al_2019}
\bibinfo{author}{Park, S.}, \bibinfo{author}{Lee, S.M.}, \bibinfo{author}{Kim,
  N.}, \bibinfo{author}{Choe, J.}, \bibinfo{author}{Cho, Y.},
  \bibinfo{author}{Do, K.H.}, \bibinfo{author}{Seo, J.B.},
  \bibinfo{year}{2019}.
\newblock \bibinfo{title}{Application of deep learning–based computer-aided
  detection system: detecting pneumothorax on chest radiograph after biopsy}.
\newblock \bibinfo{journal}{European Radiology} \bibinfo{volume}{29},
  \bibinfo{pages}{5341–5348}.
\newblock \DOIprefix\doi{10.1007/s00330-019-06130-x}.
\bibitem[{Qiu et~al.(2022)Qiu, Tan, Lin, Guan, Dai, Wang, Zhuang, Wilman,
  Huang, Cao, Tang, Jia, Li, Zhou and Wu}]{Qiu_et_al_2022}
\bibinfo{author}{Qiu, J.}, \bibinfo{author}{Tan, G.}, \bibinfo{author}{Lin,
  Y.}, \bibinfo{author}{Guan, J.}, \bibinfo{author}{Dai, Z.},
  \bibinfo{author}{Wang, F.}, \bibinfo{author}{Zhuang, C.},
  \bibinfo{author}{Wilman, A.H.}, \bibinfo{author}{Huang, H.},
  \bibinfo{author}{Cao, Z.}, \bibinfo{author}{Tang, Y.}, \bibinfo{author}{Jia,
  Y.}, \bibinfo{author}{Li, Y.}, \bibinfo{author}{Zhou, T.},
  \bibinfo{author}{Wu, R.}, \bibinfo{year}{2022}.
\newblock \bibinfo{title}{Automated detection of intracranial artery stenosis
  and occlusion in magnetic resonance angiography: A preliminary study based on
  deep learning}.
\newblock \bibinfo{journal}{Magnetic Resonance Imaging} \bibinfo{volume}{94},
  \bibinfo{pages}{105–111}.
\newblock \DOIprefix\doi{10.1016/j.mri.2022.09.006}.
\bibitem[{Redmon et~al.(2016)Redmon, Divvala, Girshick and Farhadi}]{YOLO_2016}
\bibinfo{author}{Redmon, J.}, \bibinfo{author}{Divvala, S.},
  \bibinfo{author}{Girshick, R.}, \bibinfo{author}{Farhadi, A.},
  \bibinfo{year}{2016}.
\newblock \bibinfo{title}{You only look once: Unified, real-time object
  detection}, in: \bibinfo{booktitle}{2016 IEEE Conference on Computer Vision
  and Pattern Recognition (CVPR)}, pp. \bibinfo{pages}{779--788}.
\newblock \DOIprefix\doi{10.1109/CVPR.2016.91}.
\bibitem[{Rijthoven et~al.(2018)Rijthoven, Swiderska-Chadaj, Seeliger, Laak and
  Ciompi}]{Rijthoven_et_al_2018}
\bibinfo{author}{Rijthoven, M.v.}, \bibinfo{author}{Swiderska-Chadaj, Z.},
  \bibinfo{author}{Seeliger, K.}, \bibinfo{author}{Laak, J.v.d.},
  \bibinfo{author}{Ciompi, F.}, \bibinfo{year}{2018}.
\newblock \bibinfo{title}{You only look on lymphocytes once}.
\newblock \bibinfo{journal}{Medical Imaging with Deep Learning} \URLprefix
  \url{https://openreview.net/forum?id=S10IfW2oz}.
\bibitem[{Samuelson and Petrick(2006)}]{Bootstrapping_2006}
\bibinfo{author}{Samuelson, F.}, \bibinfo{author}{Petrick, N.},
  \bibinfo{year}{2006}.
\newblock \bibinfo{title}{Comparing image detection algorithms using
  resampling}, in: \bibinfo{booktitle}{3rd IEEE International Symposium on
  Biomedical Imaging: Nano to Macro, 2006.}, pp. \bibinfo{pages}{1312--1315}.
\newblock \DOIprefix\doi{10.1109/ISBI.2006.1625167}.
\bibitem[{Savelli et~al.(2020)Savelli, Bria, Molinara, Marrocco and
  Tortorella}]{Savelli_et_al_2020}
\bibinfo{author}{Savelli, B.}, \bibinfo{author}{Bria, A.},
  \bibinfo{author}{Molinara, M.}, \bibinfo{author}{Marrocco, C.},
  \bibinfo{author}{Tortorella, F.}, \bibinfo{year}{2020}.
\newblock \bibinfo{title}{A multi-context cnn ensemble for small lesion
  detection}.
\newblock \bibinfo{journal}{Artificial Intelligence in Medicine}
  \bibinfo{volume}{103}, \bibinfo{pages}{101749}.
\newblock \DOIprefix\doi{10.1016/j.artmed.2019.101749}.
\bibitem[{Schultheiss et~al.(2020)Schultheiss, Schober, Lodde, Bodden, Aichele,
  Müller-Leisse, Renger, Pfeiffer and Pfeiffer}]{Schultheiss_et_al_2020}
\bibinfo{author}{Schultheiss, M.}, \bibinfo{author}{Schober, S.A.},
  \bibinfo{author}{Lodde, M.}, \bibinfo{author}{Bodden, J.},
  \bibinfo{author}{Aichele, J.}, \bibinfo{author}{Müller-Leisse, C.},
  \bibinfo{author}{Renger, B.}, \bibinfo{author}{Pfeiffer, F.},
  \bibinfo{author}{Pfeiffer, D.}, \bibinfo{year}{2020}.
\newblock \bibinfo{title}{A robust convolutional neural network for lung nodule
  detection in the presence of foreign bodies}.
\newblock \bibinfo{journal}{Scientific Reports} \bibinfo{volume}{10},
  \bibinfo{pages}{12987}.
\newblock \DOIprefix\doi{10.1038/s41598-020-69789-z}.
\bibitem[{Shehab et~al.(2022)Shehab, Abualigah, Shambour, Abu-Hashem, Shambour,
  Alsalibi and Gandomi}]{Shehab_et_al_2022}
\bibinfo{author}{Shehab, M.}, \bibinfo{author}{Abualigah, L.},
  \bibinfo{author}{Shambour, Q.}, \bibinfo{author}{Abu-Hashem, M.A.},
  \bibinfo{author}{Shambour, M.K.Y.}, \bibinfo{author}{Alsalibi, A.I.},
  \bibinfo{author}{Gandomi, A.H.}, \bibinfo{year}{2022}.
\newblock \bibinfo{title}{Machine learning in medical applications: A review of
  state-of-the-art methods}.
\newblock \bibinfo{journal}{Computers in Biology and Medicine}
  \bibinfo{volume}{145}, \bibinfo{pages}{105458}.
\newblock \DOIprefix\doi{https://doi.org/10.1016/j.compbiomed.2022.105458}.
\bibitem[{Shen et~al.(2017)Shen, Wu and Suk}]{Shen_et_al_2017}
\bibinfo{author}{Shen, D.}, \bibinfo{author}{Wu, G.}, \bibinfo{author}{Suk,
  H.I.}, \bibinfo{year}{2017}.
\newblock \bibinfo{title}{Deep learning in medical image analysis}.
\newblock \bibinfo{journal}{Annual Review of Biomedical Engineering}
  \bibinfo{volume}{19}, \bibinfo{pages}{221–248}.
\newblock \DOIprefix\doi{10.1146/annurev-bioeng-071516-044442}.
\bibitem[{Shen et~al.(2015)Shen, Zhou, Yang, Yang and Tian}]{Shen_et_al_2015}
\bibinfo{author}{Shen, W.}, \bibinfo{author}{Zhou, M.}, \bibinfo{author}{Yang,
  F.}, \bibinfo{author}{Yang, C.}, \bibinfo{author}{Tian, J.},
  \bibinfo{year}{2015}.
\newblock \bibinfo{title}{Multi-scale convolutional neural networks for lung
  nodule classification}, in: \bibinfo{editor}{Ourselin, S.},
  \bibinfo{editor}{Alexander, D.C.}, \bibinfo{editor}{Westin, C.F.},
  \bibinfo{editor}{Cardoso, M.J.} (Eds.), \bibinfo{booktitle}{Information
  Processing in Medical Imaging}, \bibinfo{publisher}{Springer International
  Publishing}. p. \bibinfo{pages}{588–599}.
\newblock \DOIprefix\doi{10.1007/978-3-319-19992-4_46}.
\bibitem[{Soares et~al.(2023)Soares, Castelo-Branco and
  Pinheiro}]{Soares_et_al_2023}
\bibinfo{author}{Soares, I.}, \bibinfo{author}{Castelo-Branco, M.},
  \bibinfo{author}{Pinheiro, A.}, \bibinfo{year}{2023}.
\newblock \bibinfo{title}{Microaneurysms detection in retinal images using a
  multi-scale approach}.
\newblock \bibinfo{journal}{Biomedical Signal Processing and Control}
  \bibinfo{volume}{79}, \bibinfo{pages}{104184}.
\newblock \DOIprefix\doi{https://doi.org/10.1016/j.bspc.2022.104184}.
\bibitem[{Soun et~al.(2021)Soun, Chow, Nagamine, Takhtawala, Filippi, Yu and
  Chang}]{Soun_et_al_2021}
\bibinfo{author}{Soun, J.}, \bibinfo{author}{Chow, D.},
  \bibinfo{author}{Nagamine, M.}, \bibinfo{author}{Takhtawala, R.},
  \bibinfo{author}{Filippi, C.}, \bibinfo{author}{Yu, W.},
  \bibinfo{author}{Chang, P.}, \bibinfo{year}{2021}.
\newblock \bibinfo{title}{Artificial intelligence and acute stroke imaging}.
\newblock \bibinfo{journal}{AJNR: American Journal of Neuroradiology}
  \bibinfo{volume}{42}, \bibinfo{pages}{2–11}.
\newblock \DOIprefix\doi{10.3174/ajnr.A6883}.
\bibitem[{Suzuki(2017)}]{Suzuki_2017}
\bibinfo{author}{Suzuki, K.}, \bibinfo{year}{2017}.
\newblock \bibinfo{title}{Overview of deep learning in medical imaging}.
\newblock \bibinfo{journal}{Radiological Physics and Technology}
  \bibinfo{volume}{10}, \bibinfo{pages}{257–273}.
\newblock \DOIprefix\doi{10.1007/s12194-017-0406-5}.
\bibitem[{Swiderska-Chadaj et~al.(2019)Swiderska-Chadaj, Pinckaers, Rijthoven,
  Balkenhol, Melnikova, Geessink, Manson, Sherman, Polonia, Parry, Abubakar,
  Litjens, Laak and Ciompi}]{Swiderska_et_al_2019}
\bibinfo{author}{Swiderska-Chadaj, Z.}, \bibinfo{author}{Pinckaers, H.},
  \bibinfo{author}{Rijthoven, M.v.}, \bibinfo{author}{Balkenhol, M.},
  \bibinfo{author}{Melnikova, M.}, \bibinfo{author}{Geessink, O.},
  \bibinfo{author}{Manson, Q.}, \bibinfo{author}{Sherman, M.},
  \bibinfo{author}{Polonia, A.}, \bibinfo{author}{Parry, J.},
  \bibinfo{author}{Abubakar, M.}, \bibinfo{author}{Litjens, G.},
  \bibinfo{author}{Laak, J.v.d.}, \bibinfo{author}{Ciompi, F.},
  \bibinfo{year}{2019}.
\newblock \bibinfo{title}{Learning to detect lymphocytes in
  immunohistochemistry with deep learning}.
\newblock \bibinfo{journal}{Medical Image Analysis} \bibinfo{volume}{58},
  \bibinfo{pages}{101547}.
\newblock \DOIprefix\doi{https://doi.org/10.1016/j.media.2019.101547}.
\bibitem[{Toosi et~al.(2023)Toosi, Harsini, Ahamed, Yousefirizi, Bénard, Uribe
  and Rahmim}]{Toosi_et_al_23}
\bibinfo{author}{Toosi, A.}, \bibinfo{author}{Harsini, S.},
  \bibinfo{author}{Ahamed, S.}, \bibinfo{author}{Yousefirizi, F.},
  \bibinfo{author}{Bénard, F.}, \bibinfo{author}{Uribe, C.},
  \bibinfo{author}{Rahmim, A.}, \bibinfo{year}{2023}.
\newblock \bibinfo{title}{State-of-the-art object detection algorithms for
  small lesion detection in psma pet: use of rotational maximum intensity
  projection (mip) images}, in: \bibinfo{editor}{Colliot, O.},
  \bibinfo{editor}{Išgum, I.} (Eds.), \bibinfo{booktitle}{Medical Imaging
  2023: Image Processing}, \bibinfo{publisher}{SPIE}. p.
  \bibinfo{pages}{124643E}.
\newblock \DOIprefix\doi{10.1117/12.2654527}.
\bibitem[{Tsiknakis et~al.(2021)Tsiknakis, Theodoropoulos, Manikis, Ktistakis,
  Boutsora, Berto, Scarpa, Scarpa, Fotiadis and Marias}]{Tsiknakis_et_al_2021}
\bibinfo{author}{Tsiknakis, N.}, \bibinfo{author}{Theodoropoulos, D.},
  \bibinfo{author}{Manikis, G.}, \bibinfo{author}{Ktistakis, E.},
  \bibinfo{author}{Boutsora, O.}, \bibinfo{author}{Berto, A.},
  \bibinfo{author}{Scarpa, F.}, \bibinfo{author}{Scarpa, A.},
  \bibinfo{author}{Fotiadis, D.I.}, \bibinfo{author}{Marias, K.},
  \bibinfo{year}{2021}.
\newblock \bibinfo{title}{Deep learning for diabetic retinopathy detection and
  classification based on fundus images: A review}.
\newblock \bibinfo{journal}{Computers in Biology and Medicine}
  \bibinfo{volume}{135}, \bibinfo{pages}{104599}.
\newblock \DOIprefix\doi{10.1016/j.compbiomed.2021.104599}.
\bibitem[{Wang et~al.(2022)Wang, Huang, Lee, Shen, Meng and
  Gaol}]{Wang_et_al_2022}
\bibinfo{author}{Wang, C.W.}, \bibinfo{author}{Huang, S.C.},
  \bibinfo{author}{Lee, Y.C.}, \bibinfo{author}{Shen, Y.J.},
  \bibinfo{author}{Meng, S.I.}, \bibinfo{author}{Gaol, J.L.},
  \bibinfo{year}{2022}.
\newblock \bibinfo{title}{Deep learning for bone marrow cell detection and
  classification on whole-slide images}.
\newblock \bibinfo{journal}{Medical Image Analysis} \bibinfo{volume}{75},
  \bibinfo{pages}{102270}.
\newblock \DOIprefix\doi{10.1016/j.media.2021.102270}.
\bibitem[{Wang and Yang(2018)}]{Wang_Yang_2018}
\bibinfo{author}{Wang, J.}, \bibinfo{author}{Yang, Y.}, \bibinfo{year}{2018}.
\newblock \bibinfo{title}{A context-sensitive deep learning approach for
  microcalcification detection in mammograms}.
\newblock \bibinfo{journal}{Pattern Recognition} \bibinfo{volume}{78},
  \bibinfo{pages}{12–22}.
\newblock \DOIprefix\doi{10.1016/j.patcog.2018.01.009}.
\bibitem[{Yu et~al.(2022)Yu, Wang, Hong, Teku, Wang and Zhang}]{Yu_et_al_2022}
\bibinfo{author}{Yu, X.}, \bibinfo{author}{Wang, J.}, \bibinfo{author}{Hong,
  Q.Q.}, \bibinfo{author}{Teku, R.}, \bibinfo{author}{Wang, S.H.},
  \bibinfo{author}{Zhang, Y.D.}, \bibinfo{year}{2022}.
\newblock \bibinfo{title}{Transfer learning for medical images analyses: A
  survey}.
\newblock \bibinfo{journal}{Neurocomputing} \bibinfo{volume}{489},
  \bibinfo{pages}{230–254}.
\newblock \DOIprefix\doi{10.1016/j.neucom.2021.08.159}.
\bibitem[{Yurdusev et~al.(2023)Yurdusev, Adem and Hekim}]{Yurdusev_et_al_2023}
\bibinfo{author}{Yurdusev, A.A.}, \bibinfo{author}{Adem, K.},
  \bibinfo{author}{Hekim, M.}, \bibinfo{year}{2023}.
\newblock \bibinfo{title}{Detection and classification of microcalcifications
  in mammograms images using difference filter and yolov4 deep learning model}.
\newblock \bibinfo{journal}{Biomedical Signal Processing and Control}
  \bibinfo{volume}{80}, \bibinfo{pages}{104360}.
\newblock \DOIprefix\doi{10.1016/j.bspc.2022.104360}.
\bibitem[{Zhao et~al.(2019)Zhao, Zheng, Xu and Wu}]{Zhao_et_al_2019}
\bibinfo{author}{Zhao, Z.Q.}, \bibinfo{author}{Zheng, P.}, \bibinfo{author}{Xu,
  S.T.}, \bibinfo{author}{Wu, X.}, \bibinfo{year}{2019}.
\newblock \bibinfo{title}{Object detection with deep learning: A review}.
\newblock \bibinfo{journal}{IEEE Transactions on Neural Networks and Learning
  Systems} \bibinfo{volume}{30}, \bibinfo{pages}{3212–3232}.
\newblock \DOIprefix\doi{10.1109/TNNLS.2018.2876865}.
\bibitem[{Çallı et~al.(2021)Çallı, Sogancioglu, van Ginneken, van Leeuwen
  and Murphy}]{Calli_et_all_2021}
\bibinfo{author}{Çallı, E.}, \bibinfo{author}{Sogancioglu, E.},
  \bibinfo{author}{van Ginneken, B.}, \bibinfo{author}{van Leeuwen, K.G.},
  \bibinfo{author}{Murphy, K.}, \bibinfo{year}{2021}.
\newblock \bibinfo{title}{Deep learning for chest x-ray analysis: A survey}.
\newblock \bibinfo{journal}{Medical Image Analysis} \bibinfo{volume}{72},
  \bibinfo{pages}{102125}.
\newblock \DOIprefix\doi{10.1016/j.media.2021.102125}.

\end{thebibliography}

\end{document}